\documentclass{article}

\usepackage[square,numbers,sort&compress]{natbib}

\usepackage[preprint]{corl_2026} %
\usepackage{hyperref}
\usepackage{subcaption}
\usepackage{graphicx}
\usepackage{xcolor}
\usepackage{float}
\usepackage{enumitem}
\usepackage[labelfont=bf]{caption}
\usepackage[most]{tcolorbox}
\usepackage{amssymb}
\usepackage{booktabs}
\usepackage{pdfrender}

\usepackage[nameinlink,compress]{cleveref}  %

\AtBeginEnvironment{appendices}{\crefalias{section}{appendix}}
\AtBeginEnvironment{appendices}{\crefalias{subsection}{appendix}}
\AtBeginEnvironment{appendices}{\crefalias{subsubsection}{appendix}}

\crefname{equation}{Eq.}{Eqs.}
\crefname{algorithm}{Alg.}{Algs.}
\crefname{figure}{Fig.}{Figs.}
\crefname{footnote}{Footnote}{Footnotes}
\crefname{table}{Table}{Tables}
\crefname{tabular}{Table}{Tables}
\crefname{section}{Sec.}{Sections}
\crefname{appendix}{App.}{Appendices}

\newcommand{\extrabold}[1]{%
  {\bfseries
   \pdfrender{TextRenderingMode=2,LineWidth=0.18pt}{#1}}%
}

\definecolor{navyblue}{HTML}{1f3b6f}

\newtcolorbox{contributionbox}{
    colback=navyblue!5,
    colframe=navyblue!75,
    boxrule=0.8pt,
    arc=2pt,
    left=6pt,
    right=6pt,
    top=6pt,
    bottom=6pt,
    before skip=8pt,
    after skip=8pt
}

\newtcolorbox{takeawaybox}{
    enhanced,
    colback=navyblue!5,
    colframe=navyblue!5,
    boxrule=0pt,
    borderline west={1.7pt}{0pt}{navyblue!85},
    arc=2pt,
    left=7pt,
    right=6pt,
    top=6pt,
    bottom=6pt,
    before skip=8pt,
    after skip=8pt
}

\title{Revisiting Open-Loop Execution in Robotics: 

Toward Reactive, Higher-Performing Policies}

\usepackage{color-edits}
\addauthor{mz}{red}
\addauthor{aa}{blue}
\addauthor{sa}{cyan}
\addauthor{bl}{cyan}
\addauthor{rt}{orange}

\author{
  Michael Zeng$^{1,*}$ \quad
  Abhinav Agarwal$^1$ \quad
  Ajay Bati$^3$ \quad
  Brian Lee$^1$ \quad
  Siddharth Ancha$^2$ \quad
  Russ Tedrake$^1$ \\
  $^1$Massachusetts Institute of Technology \\
  $^2$University of California, Berkeley \\
  $^3$Mundane Systems Inc. \\
  $^*$Corresponding author: \texttt{michaelszeng@gmail.com} \\[8pt]
  \href{https://revisiting-open-loop-action-chunking.github.io}
       {\texttt{https://revisiting-open-loop-action-chunking.github.io/}}
}

\begin{document}
\maketitle
\vspace{-0.1in}

\begin{abstract}
    \textit{Action chunking} --- the practice of predicting a sequence of actions and executing a prefix open-loop --- has emerged as a key enabler of recent progress in imitation learning for robotic manipulation. However, executing long open-loop prefixes reduces reactivity, limiting policies' ability to correct for errors. Further, the mechanisms underlying these performance benefits remain poorly understood: prior works cite mitigating compounding errors, absorbing inference latency, or smoothing motions, but provide limited controlled evidence or guidance for preserving reactivity. In this work, we argue that long open-loop execution primarily helps short-context policies imitate ``non-Markovian demonstrations''. Across four simulation and two real-world tasks, we show that expert non-Markovianity strongly shapes the relationship between task success and open-loop execution horizon. Further, we investigate the impact of compounding errors --- the prevailing explanation for long open-loop execution in prior work --- and find that while they matter, expert non-Markovianity has a much stronger impact in our experimental setting. Finally, we show that when policies are provided with a sufficiently long context, open-loop execution is no longer beneficial and the most reactive, closed-loop policies perform best. While imitation learning has seen great success using long open-loop execution, our findings motivate long-context, reactive policies as a more principled and performant paradigm.
\end{abstract}


\section{Introduction}

Imitation learning has recently seen rapid adoption in robotic manipulation, driven by its success on complex, contact-rich, and long-horizon tasks. A key enabler of this progress has been \textit{action chunking} --- the practice of predicting a sequence, or ``chunk," of actions and executing a fixed-length prefix open-loop before replanning from new observations. We term the length of this prefix the \textit{execution horizon}. Popularized by ACT~\cite{zhao2023learningfinegrainedbimanualmanipulation} and Diffusion Policy~\cite{chi2024diffusionpolicyvisuomotorpolicy}, long execution horizons are now used broadly across imitation learning in both academia \cite{chi2024diffusionpolicyvisuomotorpolicy, zhao2023learningfinegrainedbimanualmanipulation, kim2025finetuningvisionlanguageactionmodelsoptimizing, octomodelteam2024octoopensourcegeneralistrobot, li2025unifiedvideoactionmodel, ze20243ddiffusionpolicygeneralizable, liu2024rdt1b} and industry \cite{zhao2024alohaunleashedsimplerecipe, black2026pi0visionlanguageactionflowmodel, intelligence2025pi05visionlanguageactionmodelopenworld, figureai2025helixlogistics, nvidia2025gr00tn1openfoundation}. State-of-the-art works use open-loop execution horizons as long as 0.5 -- 1 seconds \cite{chi2024diffusionpolicyvisuomotorpolicy}.

However, using long execution horizons comes at a cost: by executing longer open-loop segments and incorporating new observations less often, it reduces policy \textit{reactivity} --- the frequency at which the policy adjusts its behavior in response to execution errors or unexpected environmental changes. We argue that reactivity has received comparatively little attention in the literature; yet, even in relatively static environments, we find uncertain dynamics and high-precision requirements make frequent re-planning desirable. Further, as manipulation advances from slow, quasi-static two-finger tasks toward fast, dexterous, and dynamic contact-rich interaction, reactivity will only become more important. Human-level dexterity will ultimately require human-level reactivity.

This motivates the question: \textbf{why do long execution horizons appear so necessary?} In particular, even at a fixed prediction horizon, we ask why longer execution horizons still improve performance. Current justifications center around reducing the effective horizon of tasks to mitigate compounding errors \cite{zhao2023learningfinegrainedbimanualmanipulation, zhang2025actionchunkingexploratorydata}, improving temporal consistency of actions to reduce unsmooth motions \cite{zhao2023learningfinegrainedbimanualmanipulation, chi2024diffusionpolicyvisuomotorpolicy}, or as a mechanism to absorb delay from policy inference or other variable latencies \cite{chi2024diffusionpolicyvisuomotorpolicy}. However, these explanations have not been evaluated through extensive, controlled experimentation. Moreover, existing justifications explain why long execution horizons are beneficial but provide little guidance on how practitioners can preserve policy reactivity.

An important fact about widely deployed imitation learning policies is that they use very short context lengths, often just the most recent 1 or 2 frames \cite{chi2024diffusionpolicyvisuomotorpolicy, trilbmteam2025carefulexaminationlargebehavior, kim2024openvlaopensourcevisionlanguageactionmodel, zhao2023learningfinegrainedbimanualmanipulation, intelligence2025pi05visionlanguageactionmodelopenworld, black2026pi0visionlanguageactionflowmodel}. This can be insufficient when behavior cloning from human demonstrations: human experts may use implicit memory and additional information unavailable to the robot when selecting actions, making them \textit{non-Markovian} with respect to the policy's observations. The central claim in this work is that \textbf{long execution horizons primarily compensate for policies' limited context lengths when trained to imitate non-Markovian experts}. We present a systematic empirical investigation of this claim across diverse tasks. We then show that increasing policy context length is an effective way to preserve reactivity while increasing task performance. Importantly, our findings do not argue against action-sequence \textit{prediction}; rather, they suggest that open-loop long-sequence \textit{execution} may not be needed.

\begin{contributionbox}
\noindent\textbf{Our key contributions are:}
\begin{enumerate}[leftmargin=2em, itemsep=2pt, topsep=4pt]
    \item We show that non-Markovianity of the expert demonstrator is a key reason long execution horizons are necessary for short-context policies.
    \item We directly compare expert non-Markovianity with compounding errors and find that, while compounding errors matter, expert non-Markovianity has a stronger influence on the optimal execution horizon than compounding errors in practice. 
    \item We show that increasing context length reduces reliance on long execution horizons; sufficiently long context removes the benefit of long execution horizons entirely; and increasing context length presents a promising direction for training highly reactive policies that achieve higher task performance.
\end{enumerate}
\end{contributionbox}

Beyond these findings, we also introduce a simple double encoder architecture that improves long-context learning, and publicly release \href{https://github.com/Michaelszeng/Revisiting-Open-Loop-Execution-in-Robotics}{code} for the automated Markovian and non-Markovian expert policies used in \cref{sec:Expert Markovianity Dominates the success-horizon curve}, supporting future studies on long-context imitation learning and the effects of expert properties.

In \cref{sec:Policies Trained on Markovian Expert Demonstrations Exhibit Optimal Execution Horizon 1}, we show that non-Markovian expert demonstrations combined with insufficient policy context length are a key reason policies benefit from long execution horizons.

In \cref{sec:Investigating the Role of Compounding Errors}, we address compounding errors as a possible explanation for using long execution horizons and compare their influence with that of expert non-Markovianity.

In \cref{sec:Long Context Provides an Alternative to Action Chunking}, we show that increasing context length can fully eliminate the benefit of long execution horizons and produce policies that are more performant than their short-context, long-execution horizon counterparts. 

In \cref{sec:Discussion}, we conclude with discussions about the role of memory and reactivity in robotic manipulation and propose future research directions.

\section{Related Works}
\label{sec:related_works}

\textbf{Action chunking in robotics:}
Action chunking (in particular, using long execution horizons) was popularized in modern robotic imitation learning by ACT \cite{zhao2023learningfinegrainedbimanualmanipulation} and Diffusion Policy \cite{chi2024diffusionpolicyvisuomotorpolicy}, which, at the time, produced breakthroughs in manipulation capabilities. Since then, using long execution horizons has become common across tabletop and mobile manipulation \cite{fu2024mobilealohalearningbimanual}, for both specialist \cite{fu2024mobilealohalearningbimanual,zhao2024alohaunleashedsimplerecipe,yu2026chi0resourceawarerobustmanipulation} and generalist robot policies \cite{trilbmteam2025carefulexaminationlargebehavior, black2026pi0visionlanguageactionflowmodel,intelligence2025pi05visionlanguageactionmodelopenworld,octomodelteam2024octoopensourcegeneralistrobot,kim2025finetuningvisionlanguageactionmodelsoptimizing,liu2024rdt1b}, and for models of all types, including flow/diffusion \cite{chi2024diffusionpolicyvisuomotorpolicy,black2026pi0visionlanguageactionflowmodel,intelligence2025pi05visionlanguageactionmodelopenworld}, variational latent-variable models \cite{zhao2023learningfinegrainedbimanualmanipulation}, autoregressive \cite{pertsch2025fastefficientactiontokenization}, and more.

Across the field, the recurring empirical pattern is that performance is often non-monotonic in the execution horizon: very short horizons underperform, intermediate horizons perform best, and very long horizons degrade performance by reducing closed-loop reactivity \cite{zhao2023learningfinegrainedbimanualmanipulation, chi2024diffusionpolicyvisuomotorpolicy}. In this work, we explore this relationship between task success and execution horizon (which we call the \textbf{\textit{success-horizon curve}}) and study the factors that determine its shape. We note that intentionally using long execution horizons is largely unique to imitation learning --- in reinforcement learning and model predictive control, for example, the highest performing robot policies employ the fastest possible closed-loop planning rate. A priori, it is unclear why short execution horizons in imitation learning perform worst.

\textbf{Current explanations for the need for open loop execution:}
A widely stated justification for using long execution horizons is that, by reducing the frequency at which the policy re-plans, long execution horizons ``reduce the effective horizon" of tasks, thus ``mitigating compounding errors" \cite{zhao2023learningfinegrainedbimanualmanipulation}. \citet{zhang2025actionchunkingexploratorydata} provide theoretical and empirical support for this explanation, showing that long execution horizons can remain beneficial even when alternative explanations such as partial observability are controlled for. We test the practical importance of compounding errors mitigation relative to expert non-Markovianity in \cref{sec:Investigating the Role of Compounding Errors}.

A practical justification for using long execution horizons is that it allows the robot to make continual progress even during latencies such as those produced by neural network inference~\cite{chi2024diffusionpolicyvisuomotorpolicy, black2025realtimeexecutionactionchunking}. Along a similar line of reasoning, many works also cite improving ``temporal action consistency" \cite{zhao2023learningfinegrainedbimanualmanipulation, chi2024diffusionpolicyvisuomotorpolicy, black2025realtimeexecutionactionchunking, malhotra2025selfguidedactiondiffusion, liu2025bidirectionaldecodingimprovingaction, yu2026chi0resourceawarerobustmanipulation, black2025trainingtimeactionconditioningefficient}, arguing that inference latency can cause newly-predicted action chunks to misalign with the previous chunk, and that increasing the execution horizon reduces the frequency at which chunk transitions occur, thus reducing unsmooth motion. While well-supported, neither explanation explains why long execution horizons are still useful in \textit{simulated experiments}, where inference latency is abstracted away due to synchronous stepping of the simulator physics and policy. In this work, we focus on understanding why long execution horizons are required even under infinitely fast (zero latency) policies.

Many works have cited elements of history-dependence or partial observability such as ``pauses in demonstrations" \cite{zhao2023learningfinegrainedbimanualmanipulation}, ``temporally correlated confounders" \cite{zhao2023learningfinegrainedbimanualmanipulation}, or ``mode commitment" \cite{chi2024diffusionpolicyvisuomotorpolicy}, which \cite{block2023provableguaranteesgenerativebehavior} unifies under the language of ``non-Markovian expert demonstrations". \cite{block2023provableguaranteesgenerativebehavior} only argues that chunked \textit{prediction}, paired with generative modeling, is necessary to capture non-Markovianity in expert demonstrations, but make no such argument about chunked \textit{execution}. Zhang et al. \cite{zhang2025actionchunkingexploratorydata} acknowledge chunked execution may be beneficial when imitating non-Markovian experts (including for imitating pauses) but focus on studying compounding errors in their work, emphasizing a result on the Robomimic Tool Hang task showing that long execution horizons remain beneficial even under a fully-observable, deterministic, Markovian expert\footnote{We find conflicting results in our experiments, but hypothesize that Zhang et al.'s result can be explained by the relatively small amount of training data used to train their learner (thus inflating compounding error), or by the low success rate ($\approx70\%$) of the expert, causing the induced expert (which filters out the actual expert's failures) to be non-Markovian. See \cref{sec:Investigating the Role of Compounding Errors}.}. We conduct experiments targeting both expert non-Markovianity and compounding errors, and conclude that expert non-Markovianity does play a strong role in the need for long execution horizons, and, in our experimental setting, usually a much stronger role than compounding errors. Concurrent work by \citet{lazzati2026doesactionchunkingimprove} similarly explains the benefits of action chunking through improved capacity to imitate non-Markovian experts, alongside reduced compounding errors and ``implicit ensembling.'' While they investigate non-Markovianity primarily through delayed policies, we study it by varying the learner's context length. These independent analyses strengthen the case that expert non-Markovianity is a key reason long execution horizons appear necessary.

\textbf{Temporal consistency and smoothness:}
Many prior works have proposed increasing temporal consistency between consecutive action chunks \cite{zhao2023learningfinegrainedbimanualmanipulation, black2025realtimeexecutionactionchunking, yu2026chi0resourceawarerobustmanipulation, black2025trainingtimeactionconditioningefficient}, with some works even claiming to reduce reliance on long execution horizons \cite{malhotra2025selfguidedactiondiffusion, liu2025bidirectionaldecodingimprovingaction}. There are a variety of ways to do so, including averaging between chunks \cite{zhao2023learningfinegrainedbimanualmanipulation,yu2026chi0resourceawarerobustmanipulation}, inference-time search \cite{liu2025bidirectionaldecodingimprovingaction}, inference-time guidance \cite{malhotra2025selfguidedactiondiffusion}, or training or inference-time inpainting \cite{black2025realtimeexecutionactionchunking, black2025trainingtimeactionconditioningefficient}. In all cases, while increasing temporal consistency sometimes allows setting shorter chunk lengths, such methods inherently reduce reactivity by binding new actions to old actions, effectively trading execution horizon for a new ``consistency" parameter. Moreover, averaging across chunks can be harmful under multimodal predictions, as it collapses distinct modes toward their mean.

\textbf{Variable chunk-length policies:}
A recent line of work treats the execution horizon itself as adaptive, either by learning execution-length selectors \cite{weng2026temporalactionselectionaction, zhao2026dynamicexecutionhorizonprediction, liang2026adaptiveactionchunkinginferencetime}, or by using test-time confidence signals to truncate chunks and replan \cite{wang2026vlaknowslimitsadaptive}. These methods can improve performance by trading off the shortcomings of long and short execution horizons at favorable times, but do not address why short-horizon policies fail in the first place. We target this root cause directly.

Another set of works trains low-frequency chunking base policies with high-frequency reactive heads \cite{niu2026trextactilereactivedexterousmanipulation, xue2025reactivediffusionpolicyslowfast, sendai2025leaveobservationbehindrealtime, wu2026closedloopactionchunksdynamic, wang2026phaforcephasescheduledvisualforcepolicy}. However, the reactive heads typically condition on a smaller set of observations and are bound to the outputs of the low-frequency base policies (i.e. by conditioning on latent action chunks \cite{xue2025reactivediffusionpolicyslowfast} or using a partially denoised output from the base policy \cite{niu2026trextactilereactivedexterousmanipulation}), reducing them to only local refinement. We seek reactivity to even large environment changes that require full-policy re-planning.

\textbf{Long-context imitation learning:} As our findings show that context length significantly affects the optimal execution horizon, we review relevant works in the space of long-context imitation learning.
Long context lengths are beneficial for imitation learning policies to capture history-dependencies present in expert data or required by the task, but a vast majority of imitation-learning policies are memory-less or condition on only a short history (of 1 -- 2 frames) due to the \textit{causal confusion} problem in long-context learning \cite{dehaan2019causalconfusionimitationlearning}. Recent work has pursued long-context imitation learning through two broad directions: directly scaling raw temporal context \cite{torne2025learninglongcontextdiffusionpolicies, agarwal2026trainingevaluatingdiffusionpolicies, torne2026memmultiscaleembodiedmemory}, and selectively compressing, retrieving, or maintaining task-relevant memory \cite{torne2026memmultiscaleembodiedmemory, mark2026bpplongcontextrobotimitation, shah2026memoryretrievalvisuomotorpoliciesHALO, sridhar2025memerscalingmemoryrobot, lei2026vpwemnonmarkovianvisuomotorpolicy, guo2026chameleoncontrolindexedprospective, zheng2025tracevlavisualtraceprompting, shi2026memoryvlaperceptualcognitivememoryvisionlanguageaction, li2025mapvlamemoryaugmentedpromptingvisionlanguageaction, lin2026echovlasynergisticdeclarativememory, lin2026hifvlahindsightinsightforesight}. For our experiments, we are interested in the first line of works because they retain maximum observability for inferring expert hidden state, which is important when imitating non-Markovian expert demonstrations.

\section{Preliminaries}

\subsection{Imitation learning and Diffusion Policy}

In behavior cloning, successful task demonstrations are collected from an expert and used as supervised data to train a policy that reproduces the expert’s actions. We represent these demonstrations as a dataset of observation-action pairs, $\mathcal D = {(O_t^{(i)}, A_t^{(i)})}_{i=1}^n$, and train a policy $\pi_\theta(A_t \mid O_t)$ to approximate the expert conditional distribution $p_E(A_t \mid O_t)$. Diffusion Policies \cite{chi2024diffusionpolicyvisuomotorpolicy} accomplish this by using a diffusion model to learn the relevant conditional distribution; we provide additional background on their training and inference in \cref{sec:appendix_difffusion_policy}.

Following \cite{chi2024diffusionpolicyvisuomotorpolicy, zhao2023learningfinegrainedbimanualmanipulation}, modern behavior cloning methods learn to predict a ``chunk" of actions $A_t = (a_t, a_{t+1}, \ldots, a_{t+T_p-1})$ given a short history of observations: $O_t = (o_{t-T_o+1}, \ldots, o_t)$ where $T_o$ denotes the observation horizon or \textbf{context length} and $T_p$ denotes the \textbf{prediction horizon}. A vast majority of works use short context lengths of $T_o = 1$ or $2$ \cite{zhao2023learningfinegrainedbimanualmanipulation,chi2024diffusionpolicyvisuomotorpolicy,octomodelteam2024octoopensourcegeneralistrobot,liu2024rdt1b,black2026pi0visionlanguageactionflowmodel,kim2025finetuningvisionlanguageactionmodelsoptimizing,ze20243ddiffusionpolicygeneralizable}. 

These policies typically execute a prefix of the predicted action chunk open-loop before re-querying the policy on new observations. We let $T_{\text{exec}}$ denote the \textbf{execution horizon}: the number of predicted actions executed before re-querying the policy. We use $T_{\text{exec}}^*$ for the value of $T_{\text{exec}}$ that maximizes task performance.

\textbf{This work studies the effect of $T_{\text{exec}}$ on policy performance given a fixed $T_p$.} In particular, while $T_{\text{exec}}=1$ is necessary for closed-loop reactive control, we study why $T_{\text{exec}} > 1$ has generally been deemed necessary for behavior cloning to succeed.

Strictly speaking, the relevant quantity is the open-loop \textit{execution time} between successive policy queries. However, because demonstrations are usually discretized with a fixed time period $dt$ between consecutive actions, this duration is \mbox{$T_{\text{exec}} \cdot dt$}. In this work, we fix $dt$ and only vary $T_{\text{exec}}$ and treat $T_{\text{exec}}$ synonymously with execution time.

Throughout our experiments, we use the term \textbf{\textit{success-horizon curve}} to denote the relationship between policy success rate and execution horizon ($T_{\text{exec}}$), plotted on a 2D curve, with task, dataset, policy class, and training procedure held constant. See \cref{fig:human_vs_markovian_vs_non_markovian_furnituresim} for an example.

\subsection{Compounding Errors in Imitation Learning}
Imitation learning policies are trained on states visited by the expert, but at test time, small policy errors can move the robot into states that are unlikely under the expert demonstrations. The policy is then forced to act on out-of-distribution observations, which can lead to further errors that compound over time~\citep{ross2011reductionimitationlearningstructured}. This phenomenon is known as \textit{compounding errors} and is often used to motivate long execution horizons: executing more actions per policy query reduces the effective closed-loop decision horizon~\citep{zhao2023learningfinegrainedbimanualmanipulation, simchowitz2025pitfallsimitationlearningactions} and, consequently, the severity of compounding errors.

Data aggregation (DAgger~\citep{ross2011reductionimitationlearningstructured}) is a standard method for reducing compounding errors: it iteratively rolls out the learned policy, queries the expert for the correct actions on states actually visited by that policy, aggregates these newly labeled states into the training dataset, and re-trains, thereby reducing mismatch between the training state distribution and policy-induced test-time state distribution. 

We apply a variant of DAgger called HG-DAgger in \cref{sec:Investigating the Role of Compounding Errors} as an intervention to test whether long execution horizons primarily help by mitigating compounding errors.

\subsection{Non-Markovian experts}
We define a setup similar to \citet{block2023provableguaranteesgenerativebehavior}. For simplicity, we treat the environment observations $o_t$ as Markovian states for the environment.\footnote{This approximation is commonly used for algorithmic discussion in robotic behavior cloning \cite{intelligence2025pi06vlalearnsexperience}. While not exactly true, it is often reasonable with sufficient camera views.} The robot accomplishes the task in an environment with Markovian dynamics. That is, given $o_t$ and $a_t$, the environment dynamics evolve as
$p(o_{t+1} \mid o_{\leq t}, a_{\leq t}) = p(o_{t+1} \mid o_t, a_t)$.
We define the following two forms for the expert $E$ providing demonstrations in this environment: 
\begin{itemize}[leftmargin=2.5em]
    \item \textbf{Markovian Expert:} The expert’s action distribution depends only on the current environment state:  $\pi_E(a_t | o_{\leq t}) = \pi_E(a_t | o_t)$. 
    \item \textbf{Non-Markovian Expert:} The expert’s action distribution depends on both the current and past environment states: $\pi_E(a_t \mid o_{\leq t}) \neq \pi_E(a_t \mid o_t)$. We qualify experts which maintain a hidden state that carries distilled information from past observations in this category. For instance, experts providing demonstrations which require them to keep track of time.
\end{itemize}

Notably, human experts are inevitably non-Markovian: even when trying not to, they naturally draw on earlier actions and observations, latent preferences, and other unrecorded information when providing demonstrations.

\section{Expert Markovianity Dominates the Success-Horizon Curve}
\label{sec:Expert Markovianity Dominates the success-horizon curve}
We now demonstrate our first result: \textbf{the success-horizon curve is strongly influenced by Markovianity of the expert when training short-context policies.}

We compare the success-horizon curves of Diffusion Policies trained on Markovian and non-Markovian expert data across 4 diverse simulation benchmarks.
\emph{When the policy architecture, training procedure, and evaluation setting are fixed, we find that changing only the nature of the expert to Markovian shifts the curve toward near-monotonically decreasing with optimal execution horizon $T_{\text{exec}}^*=1$}. We then address the compounding errors hypothesis --- a common justification for long execution horizons from prior work --- directly and show that, on the \texttt{FurnitureSimOneLeg} task \cite{heo2023furniturebenchreproduciblerealworldbenchmark}, interventions designed to reduce compounding errors have a much smaller effect on the success-horizon curve than changing expert Markovianity.

\begin{figure}[h!]
    \centering
    \includegraphics[width=1\linewidth]{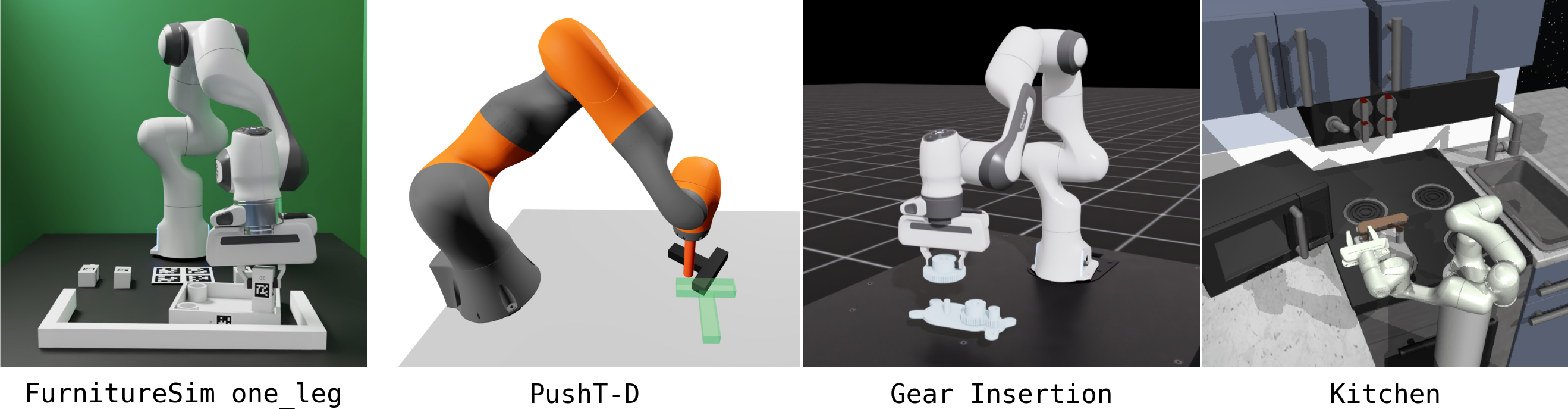}
    \caption{\textbf{Benchmark Tasks.} \textbf{{\texttt{FurnitureSimOneLeg}}} \cite{heo2023furniturebenchreproduciblerealworldbenchmark}: pick up table leg and insert and screw into table top to assemble the table. \textbf{{\texttt{Push-T}}} \cite{wei2025empiricalanalysissimandrealcotraining}: Push T-shaped slider to target pose, then return robot to reset pose. \textbf{{\texttt{GearInsertion}}}: Pick up large gear and insert onto gear carrier, meshed with other gears. \textbf{{\texttt{Kitchen}}} \cite{gupta2019relaypolicylearningsolving}: Complete any 4 kitchen subtasks (e.g., open microwave, move kettle, turn top/bottom knob, flip switch, open slide door, open hinge door), in any order; the dataset contains random subtask permutations. \textbf{All tasks use image-based observations.}}
    \label{fig:tasks}
\end{figure}

In all experiments in this section, we train image-based Diffusion Policies that use context length $T_o=2$, prediction horizon $T_p=15$, and execution horizons $T_{\text{exec}}$ ranging from $1$ to $15$, holding all training and evaluation parameters constant besides the dataset.\footnote{We make an exception for \texttt{Kitchen}; we additionally provide a subtask sequence label to ensure full observability. See \cref{sec:appendix_a_markovian_experts}.} We choose a variety of types of benchmarks, ranging from high-precision (\texttt{FurnitureSimOneLeg}, \texttt{GearInsertion}) to highly multimodal (\texttt{Kitchen}) and contact-rich (\texttt{Push-T}). See \cref{fig:tasks} for task descriptions. \textbf{Importantly, we conduct experiments in the moderate-to-high data regime, training policies on 200+ demonstrations, depending on the task. We believe experiments in this regime are important to ensure results transfer to deployment systems, which are increasingly trained on large datasets.}

For \texttt{Push-T}, we use the ManiSkill simulator \cite{mu2021maniskillgeneralizablemanipulationskill} for the expert comparisons in \cref{sec:Policies Trained on Markovian Expert Demonstrations Exhibit Optimal Execution Horizon 1} (\texttt{PushT-M}) and Drake \cite{tedrake2019drake} for the context-length experiments in \cref{sec:Long Context Provides an Alternative to Action Chunking} (\texttt{PushT-D}). We distinguish them because simulator physics can materially affect planar pushing: Drake provides richer hydroelastic contact \cite{masterjohn2022velocitylevelapproximationpressure}, while ManiSkill’s GPU parallelism makes training a Markovian RL expert more practical. See \cref{sec:appendix_a_benchmark_task_details} for details.

\subsection{Policies Trained on Markovian Expert Demonstrations Exhibit Optimal Execution Horizon $T_{\text{exec}}^*=1$}
\label{sec:Policies Trained on Markovian Expert Demonstrations Exhibit Optimal Execution Horizon 1}

Firstly, we isolate the effect of expert Markovianity by testing whether replacing non-Markovian demonstrations with Markovian demonstrations shifts the optimal execution horizon to $T_{\text{exec}}^*=1$.

\subsubsection{Experimental Setup} 
\label{sec:markovian_experts_experimental_setup}

We develop two scripted experts for the \texttt{FurnitureSimOneLeg} task. 
\begin{enumerate}[leftmargin=2.5em]
    \item \textbf{Deterministic Markovian Expert:} This expert is implemented as a finite-state machine (FSM) that determines its state using the current and previous environment states, and uses end-effector operational-space control (OSC) and gripper PID control to move the robot toward FSM state-specific waypoints. Note that the Diffusion Policy conditions on 2 frames of history ($T_o=2$), so a 2-Markovian expert (that conditions on current and previous environment states) is Markovian with respect to the policy's input.
    \item \textbf{Non-Markovian Expert:} This expert is constructed by injecting a variety of non-Markovian behaviors into the deterministic Markovian expert, including a hidden latent plan (consisting of episodically pre-determined waypoint offsets), sticky FSM state transitions (requiring the expert to stay in the current state for a latent number of steps (up to 16) before advancing), and latent-count alignment maneuvers (repeatedly noising the trajectory before picking or inserting the leg, for up to 10 timesteps per iteration and 3 iterations). These injections are designed to mimic human behaviors (e.g. pauses, delay in mentally registering subtask completion, sub-optimal or cyclic alignment motions).
\end{enumerate}

In addition to these two scripted experts, we also compare a \textbf{human teleoperator} as an inherently non-Markovian expert. 

\begin{figure}[h!]
    \centering
    \includegraphics[width=1\linewidth]{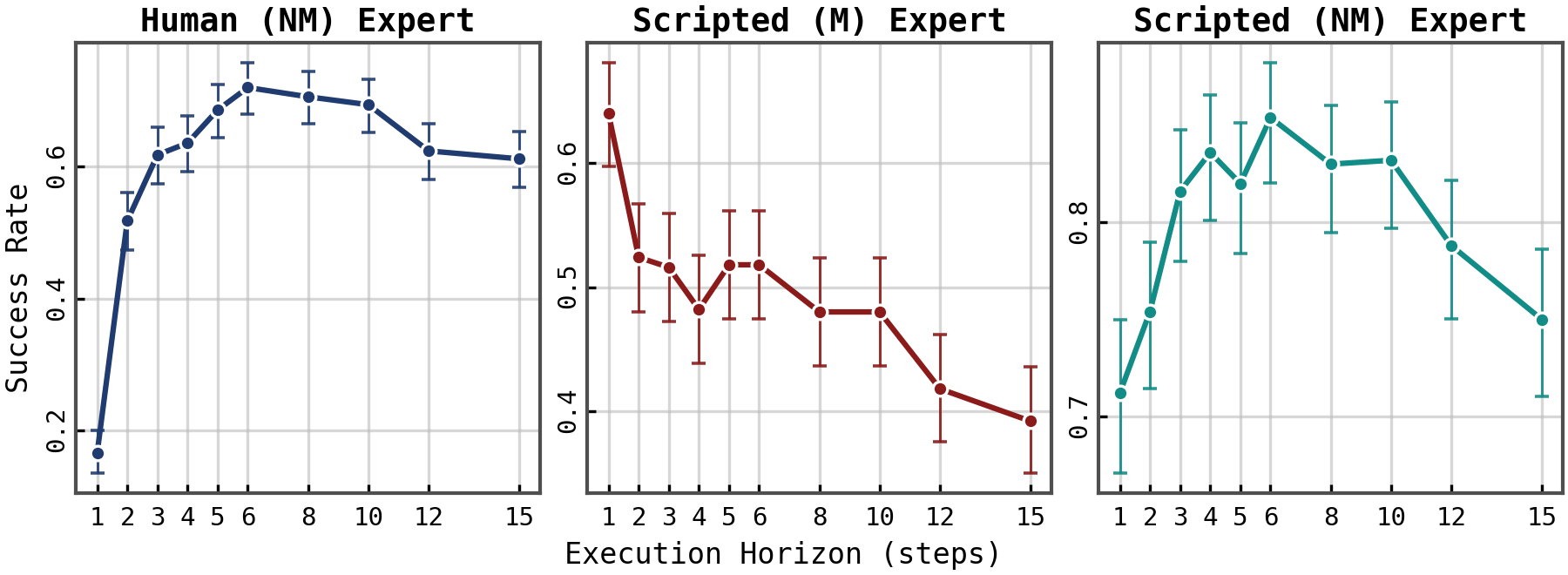}
    \caption{\textbf{Success-horizon curves for Furniture-Sim \texttt{one-leg}.} The policy trained on human non-Markovian (NM) expert data (\textbf{Left}) exhibits the typical inverted U-shape curve; the policy trained on Markovian (M) scripted expert data (\textbf{Center}) exhibits an approximately monotonically decreasing curve; adding non-Markovian behaviors to the scripted expert (\textbf{Right}) restores the inverted U-shape.}
    \label{fig:human_vs_markovian_vs_non_markovian_furnituresim}
\end{figure}

\begin{figure}[h!]
    \centering
    \includegraphics[width=1\linewidth]{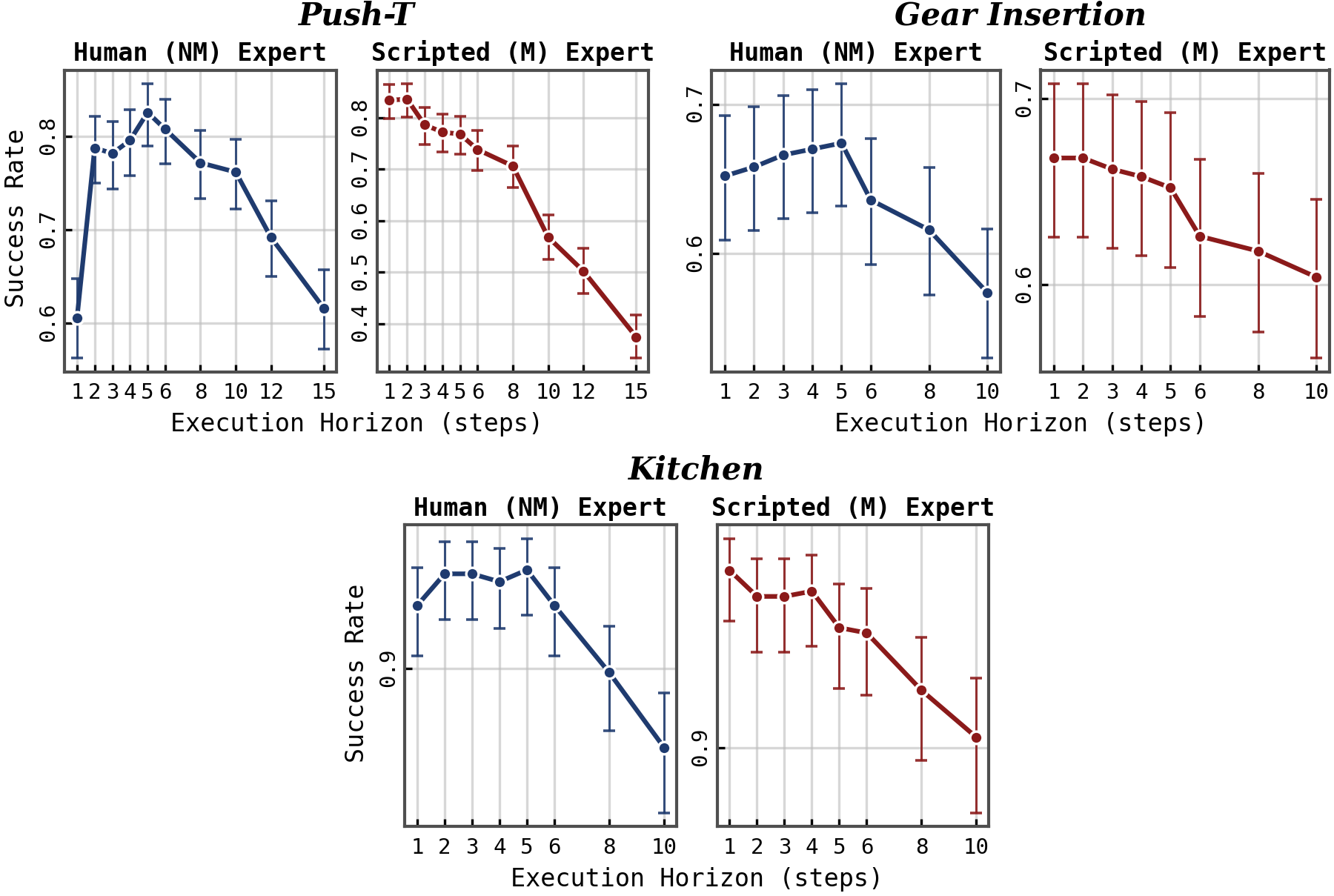}
    \caption{\textbf{Success-horizon curve with non-Markovian (NM) human vs scripted Markovian (M) expert demonstrations for \texttt{PushT-M} (\textbf{Top Left}), \texttt{GearInsertion} (\textbf{Top Right}), and \texttt{Kitchen} (\textbf{Bottom}) tasks.} All exhibit the typical inverted U-shaped curve when trained on human data, but all become monotonically decreasing when trained on Markovian expert data.}
    \label{fig:human_vs_markovian_3_tasks}
\end{figure}

We evaluate each expert at a variety of execution horizons and plot results in \cref{fig:human_vs_markovian_vs_non_markovian_furnituresim}.
We repeat the experiment for \texttt{PushT-M}, \texttt{GearInsertion}, and \texttt{Kitchen} in \cref{fig:human_vs_markovian_3_tasks}, plotting the success-horizon curve under \textbf{Markovian} and \textbf{human} experts. Further experimental details are in \cref{sec:appendix_a_markovian_experts}.

\subsubsection{Results and Discussion}
\label{subsec:main_expert_nature_results}
\textbf{Expert Markovianity changes the success-horizon curve to monotonically decreasing:} All policies trained on human expert data exhibit the inverted U-shape curve as commonly seen in the literature~\cite{zhao2023learningfinegrainedbimanualmanipulation,chi2024diffusionpolicyvisuomotorpolicy}. All policies trained on Markovian expert data exhibit optimal execution horizon of $T_{\text{exec}}^* = 1$ and near-monotonic decrease in success rate with $T_{\text{exec}}$. Given that the expert demonstrator was the only factor varied between the two sets of experiments, we determine that the nature of the Markovian expert's data is the driver of the shift in success-horizon curve in each benchmark task. 

Notably, for \texttt{FurnitureSimOneLeg}, simply injecting non-Markovian behaviors into the Markovian scripted expert restores the inverted U-shape, showing that certain non-Markovian behaviors necessitate long execution horizons. In particular, we find non-Markovian features that make expert behavior depend on information especially difficult for the policy to infer from its 2 frames of context to be most detrimental, such as sticky FSM transitions or latent-count alignment maneuvers, where the policy is effectively unable to observe the passage of time or the latent count (for example, while imitating a long pause, the policy is unable to determine how long it has been paused for based on 2 frames of context).

\textbf{Low-horizon failures are consistent with our non-Markovianity hypothesis:} Observing policy performance qualitatively, we find that dominant failure modes are caused by ambiguity in the desired action distribution when re-planning, caused by the expert being non-Markovian. With only a context length of 2, policies may select actions that appear valid based on their limited context, but are actually inconsistent with what the full-history conditioned expert would select. This can lead to mode-switching or unbounded repetition of locally valid motions.

\begin{figure*}[h!]
    \centering
    \subcaptionbox{\label{fig:failure_case_furnituresim_insertion}}{%
        \includegraphics[height=3.12cm]{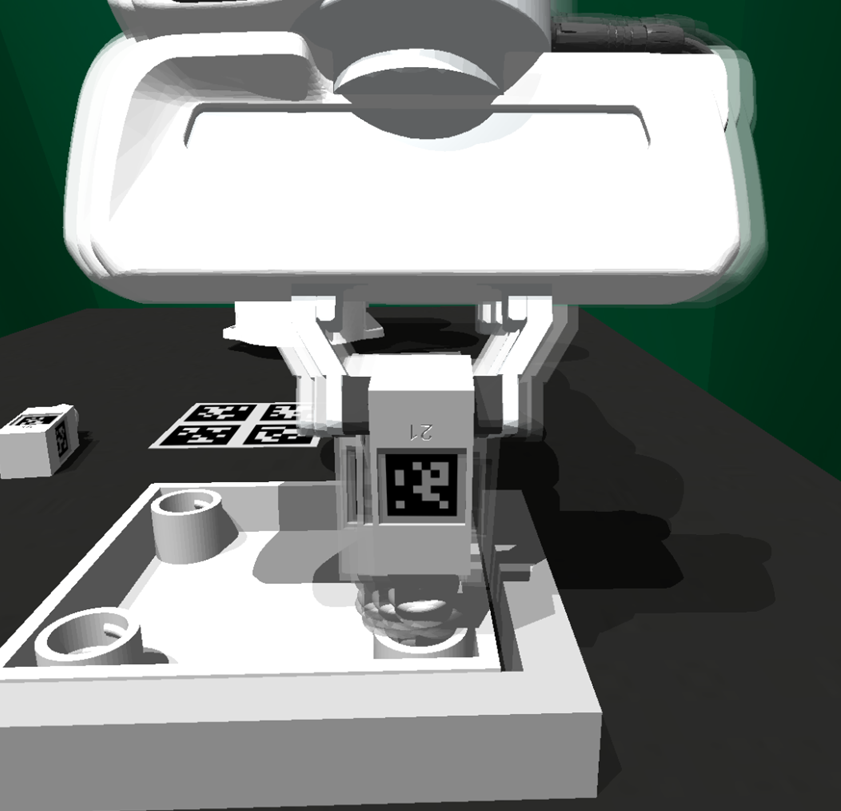}%
    }%
    \hspace{0.2mm}%
    \subcaptionbox{\label{fig:failure_case_furnituresim_pick}}{%
        \includegraphics[height=3.12cm]{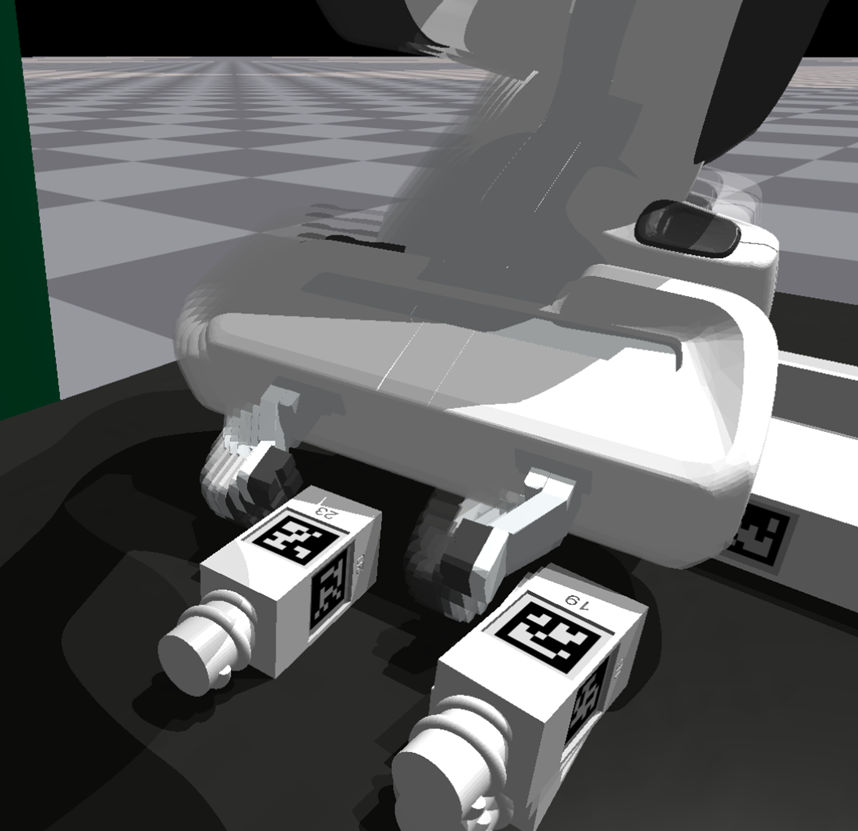}%
    }%
    \hspace{0.2mm}%
    \subcaptionbox{\label{fig:failure_case_gear_insertion}}{%
        \includegraphics[height=3.12cm]{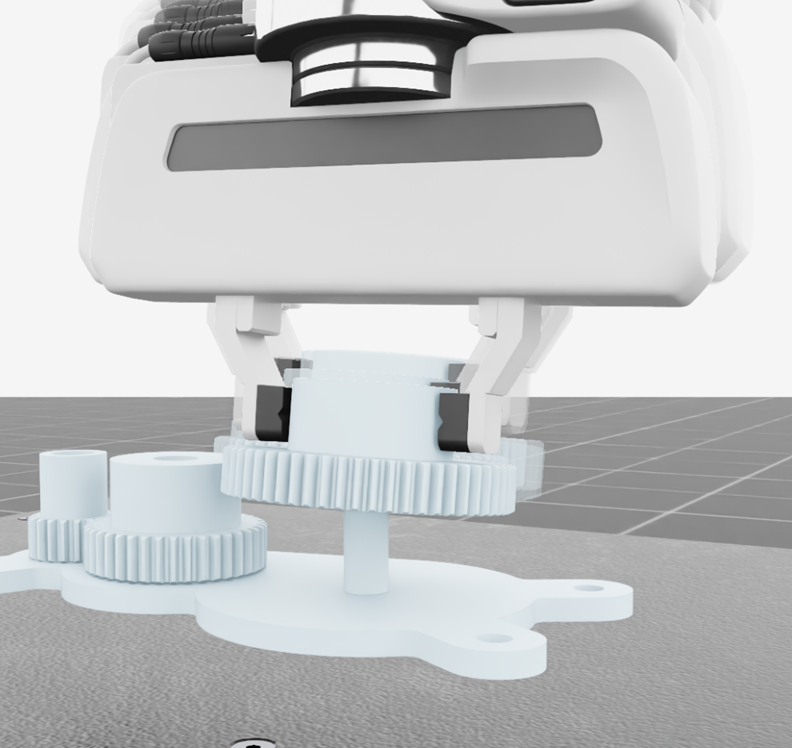}%
    }%
    \hspace{0.2mm}%
    \subcaptionbox{\label{fig:failure_case_push_t}}{%
        \includegraphics[height=3.12cm]{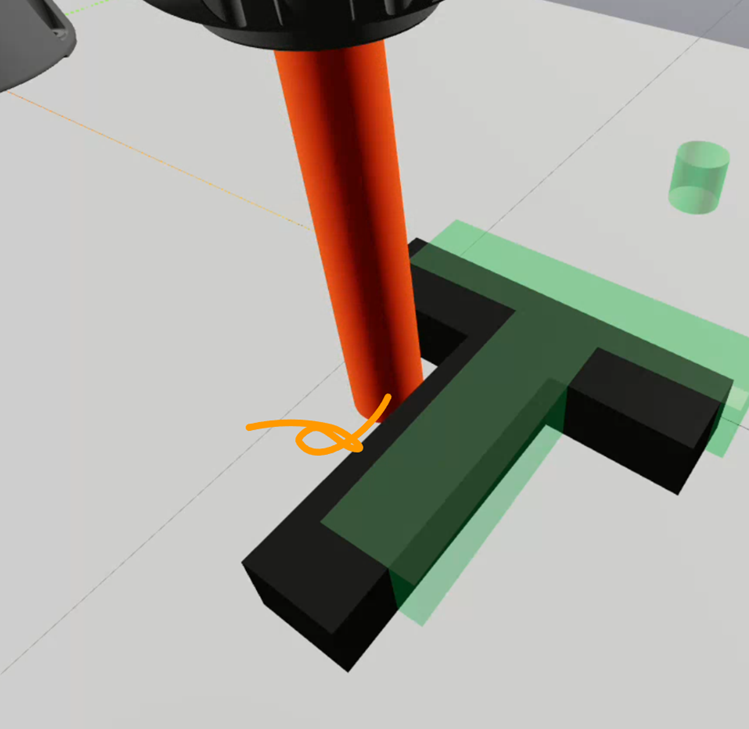}%
    }
    \hspace{0.2mm}%
    \subcaptionbox{\label{fig:failure_case_kitchen}}{%
        \includegraphics[height=3.12cm]{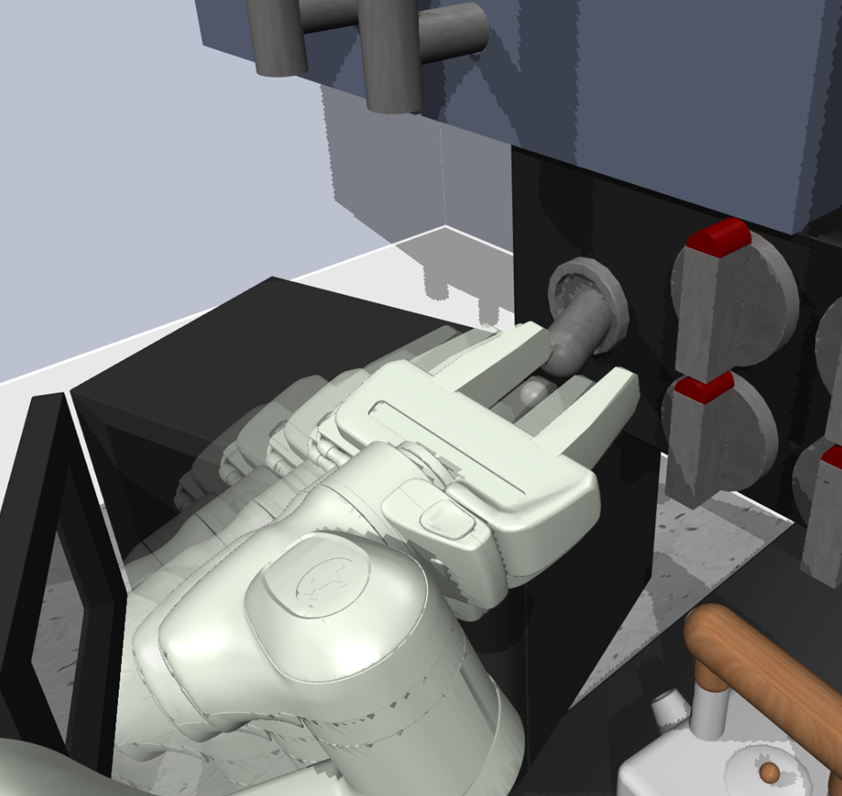}%
    }%
    \caption{\textbf{Illustration of common failure modes of short-context policies using short execution horizons}. \textbf{(a)}: \texttt{FurnitureSimOneLeg}: robot cycles between coherent alignment motions but fails to transition to insertion, causing timeout. \textbf{(b)}: \texttt{FurnitureSimOneLeg}: robot performs repeated alignment motions with the table leg, but fails to transition to grasp. \textbf{(c)}: \texttt{GearInsertion}: robot repeatedly aligns gear with shaft. \textbf{(d)}: \texttt{Push-T}: the robot jitters until timeout, oscillating between two modes (pushing the T rightward versus moving around to push it upward). \textbf{(e)}: \texttt{Kitchen}: robot approaches light switch but abruptly changes course toward the top knob and collides.}
    \label{fig:failure_cases}
\end{figure*}

For instance, in \texttt{FurnitureSimOneLeg}, we observe many failures around aligning to pick up the table leg or aligning before insertion; the robot hovers and repeats valid approach maneuvers as demonstrated by the expert, but never commits to the pick or insertion. In \texttt{GearInsertion}, similar failures occur when the robot aligns the gear to the shaft, rapidly switching between behaviors until failure. In \texttt{PushT-M}, because of how multimodal the task and demonstrations are, repeated mode-switching occurs almost anywhere around the T. In \texttt{Kitchen}, we observe the policy switch from one expert-consistent motion to another when very near two subtasks, ultimately completing neither. We visualize these failure modes in \cref{fig:failure_cases}. Taken together, these behaviors point to expert non-Markovianity as the root cause: short-context policies cannot infer which of several locally valid expert behaviors is consistent with the entire trajectory history. Long execution horizons reduce this planning ambiguity by simply committing to a single mode for longer and re-planning less often.

We discuss this mechanism further in \cref{sec:Why Long Execution Horizons Prevent Failure}.

So far we have only commented on how long execution horizons help address multimodality arising from the policy having limited context compared to the expert. However, the action distribution induced by a collection of expert demonstrations might itself be multimodal and thus could plausibly cause mode-switching in policy execution as well; while we don't explicitly study this case here, we discuss it in  \cref{appx:multimodality_and_smooth_execution}.

\textbf{Reactivity is consistently beneficial even in quasi-static tasks:} We note that none of our tasks are inherently dynamic or unpredictable. Nevertheless, across all tasks, policies trained on Markovian expert data perform best at or near $T_{\text{exec}}=1$, when they are most reactive. These experiments show that, even in seemingly static tasks, errors in execution and in the learned dynamics are nearly unavoidable, making reactive feedback broadly beneficial in manipulation, not only in inherently dynamic settings. While ablating $T_{\text{exec}}$ is not common in prior work, selecting the best $T_{\text{exec}}$ can yield substantial improvements of tens of percentage points in absolute success rates. We discuss concrete examples of the benefits of reactivity in these benchmark tasks in \cref{sec:appendix_reactivity}.

\begin{takeawaybox}
\textbf{Key Takeaway.} Across tasks, longer execution horizons are beneficial when short-context policies imitate non-Markovian demonstrations. When the expert is Markovian, the optimal execution horizon shifts to $T_{\text{exec}}^*=1$, and the most reactive policies perform best. Expert non-Markovianity is therefore a key driver of the success-horizon curve.
\end{takeawaybox}

\subsection{Investigating the Role of Compounding Errors}
\label{sec:Investigating the Role of Compounding Errors}

\citet{zhang2025actionchunkingexploratorydata} identify mitigating compounding errors as a key function of long execution horizons. We test the practical importance of this mechanism on \texttt{FurnitureSimOneLeg} using two interventions designed to improve train–test state-distribution alignment and compare their effects with those of changing expert Markovianity. If compounding errors strongly determine the success-horizon curve, these interventions should disproportionately improve performance with short execution horizons and shift the optimum toward $T_{\text{exec}}^*=1$. Across these interventions, we observe only modest shifts compared with those caused by changing expert Markovianity, suggesting that compounding errors matter but do not alone explain the success-horizon curve in this setting.

\textbf{Intervention 1: Data scale.} We begin with the simplest intervention: increasing the amount of training data. While this does not directly correct mismatch between the expert and policy-induced state distributions, broader state coverage is more likely to include states induced by the learned policy, which can mitigate compounding errors~\cite{laskey2017dartnoiseinjectionrobust,chang2022mitigatingcovariateshiftimitation}. 

\begin{figure}[h!]
    \centering
    \includegraphics[width=0.95\linewidth]{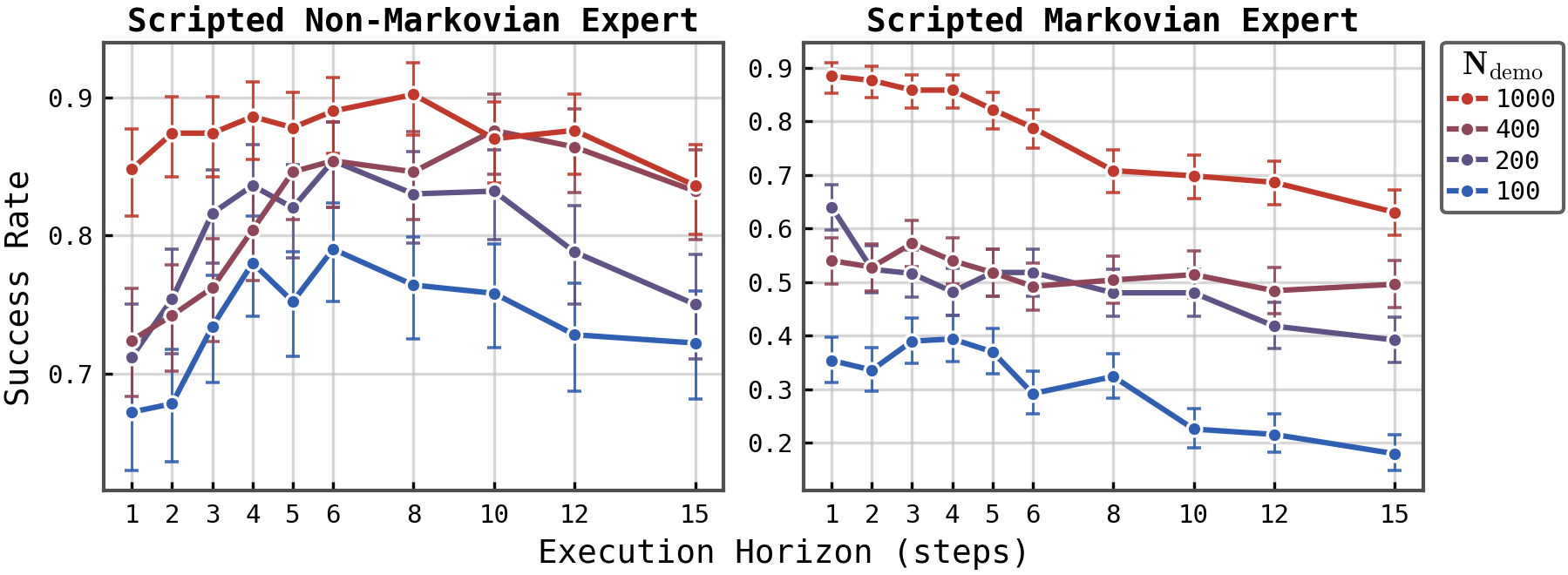}
    \caption{\textbf{Success-horizon curve at dataset sizes $\in \{100, 200, 400, 1000\}$ demonstrations}, using both Markovian (\textbf{Right}) and non-Markovian (\textbf{Left}) scripted expert data. Only at the lowest data scale (100 demonstrations), we find compounding errors to play a significant role in shaping the curve, causing even the policy trained on Markovian expert data to prefer a long execution horizon. Above 200 demonstrations, we find increasing dataset size only modestly shifts preference toward lower execution horizons.}
    \label{fig:data_ablation}
\end{figure}

In \cref{fig:data_ablation}, for the scripted Markovian expert, we observe a clear distinction between the lowest data scale and the moderate-to-high data regime. With only 100 demonstrations, even the policy trained on deterministic Markovian expert data performs best at a non-trivial execution horizon of $T_{\text{exec}}=4$, indicating that limited state coverage and the resulting compounding errors can substantially shape the success-horizon curve. From 200 to 1,000 demonstrations, however, additional data produces only a slight tilt toward shorter execution horizons, indicating that compounding errors have comparatively modest influence once moderate state coverage is reached. Thus, our results support Zhang et al.'s hypothesis in the most data-limited setting, but suggest that, in the moderate-to-high data regime targeted by our main experiments, compounding errors are a secondary contributor with a substantially smaller effect than expert non-Markovianity.

\textbf{Intervention 2: HG-DAgger.} Next, we run 3 iterations of semi-automated human-gated DAgger (HG-DAgger) \cite{kelly2019hgdaggerinteractiveimitationlearning} on our policies trained on 200 Markovian and non-Markovian expert demonstrations. For each execution horizon independently, we first roll out the base policy, record failed episodes, manually annotate an intervention timestep for each failure, reset the simulation to that timestep, and collect an expert correction from that timestep onward. We then fine-tune the policy on the additional correction data, and repeat this process for 3 iterations. In our experiment, each round of HG-DAgger adds 40 correction demonstrations. Note that we only consider $T_{\text{exec}} \in \{1,3,6,10,15\}$ instead of the denser sweep used in our other experiments because of the large amount of manual annotation required per value of $T_{\text{exec}}$. See \cref{sec:appendix_hg_dagger} for more details. 

\begin{figure}[h!]
    \centering
    \includegraphics[width=0.95\linewidth]{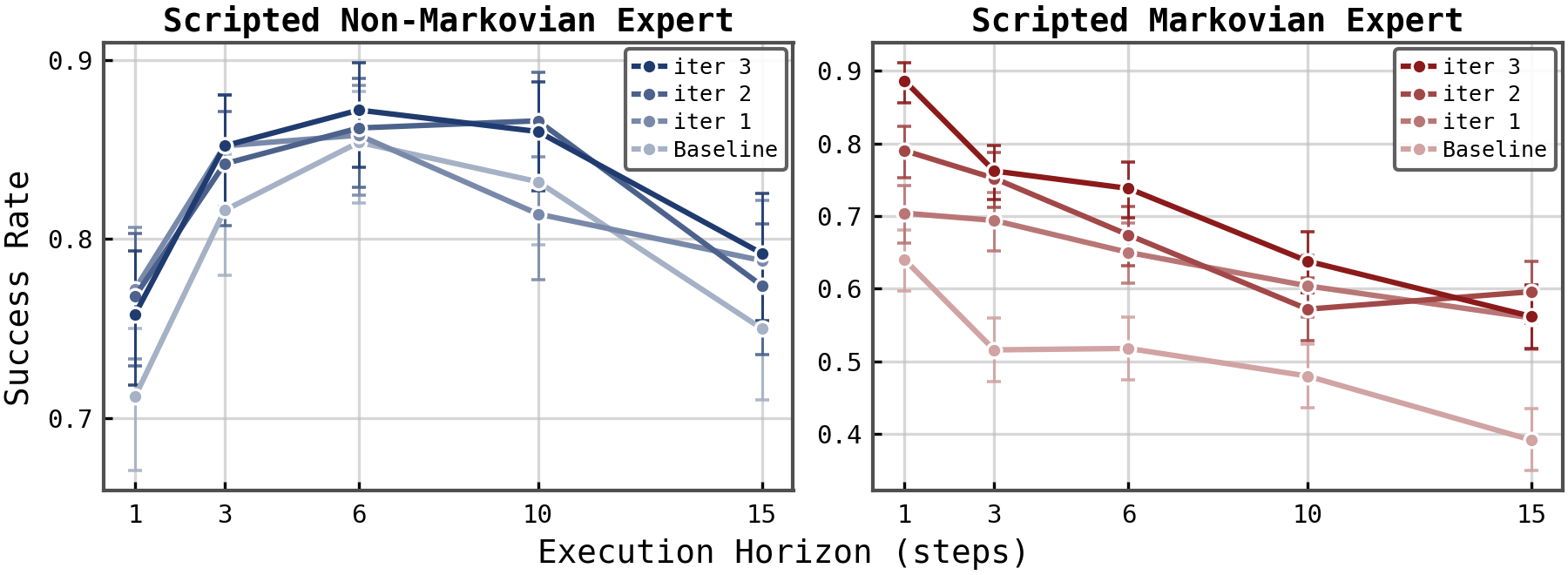}
    \caption{\textbf{Success-horizon curves after applying HG-DAgger} for Scripted Markovian (\textbf{Right}) and non-Markovian (\textbf{Left}) expert-trained policies. We find no noticeable shift for the non-Markovian expert-trained policy, and a slight shift favoring lower execution horizons for the Markovian expert-trained policy.}
    \label{fig:DAgger}
\end{figure}

In \cref{fig:DAgger}, we find no significant shift over the 3 HG-DAgger iterations for the policy trained on non-Markovian expert data, and a slight shift favoring lower execution horizons for the policy trained on Markovian expert data. These results support that compounding errors contribute to the success-horizon curve for the Markovian expert case, but the effects become less noticeable in the presence of expert non-Markoviany. Qualitatively, we observe that successive HG-DAgger rounds primarily correct basic manipulation failures, such as missed insertions and screwing failures. For policies trained on non-Markovian expert data, however, HG-DAgger does not eliminate failures caused by cyclic or idle behaviors, such as repeatedly aligning to the leg during pick-up or repeatedly aligning before insertion (see \cref{fig:failure_case_furnituresim_insertion,fig:failure_case_furnituresim_pick}). Because the corrections remain non-Markovian, HG-DAgger does not remove (but rather perhaps reinforces) the ambiguity faced by short-context policies when re-planning that causes these cyclic failures.

Finally, we provide several supporting analyses in the appendix. We evaluate exploratory noise injection, a complementary compounding errors mitigation technique proposed by \citet{zhang2025actionchunkingexploratorydata}. We also examine how proper checkpoint selection and policy architecture both demonstrate the impact of compounding errors while showing how it can be limited in practice. See \cref{sec:appendix_noise_injection} (exploratory noise injection), \cref{sec:appendix_checkpoint_selection} (checkpoint selection), and \cref{sec:pusht-m-additional-discussion} (architectural effects).

\begin{takeawaybox}
\textbf{Key Takeaway.}

Across interventions designed to mitigate compounding errors: (i) increasing dataset size and (ii) HG-DAgger, we observe at most modest shifts toward shorter execution horizons, even when the interventions improve overall success. Within \texttt{FurnitureSimOneLeg} and the interventions tested, expert Markovianity therefore appears to have a stronger influence on the success-horizon curve than compounding errors.
\end{takeawaybox}

Given that results in \Cref{subsec:main_expert_nature_results} are consistent across tasks, we expect the results in this section to readily extend to other tasks as well. 

\section{Long Context Reduces the Need for Long Execution Horizons}
\label{sec:Long Context Provides an Alternative to Action Chunking}

\textbf{In this section we demonstrate that training policies with longer context lengths reduces reliance on long execution horizons and enables higher-performing \textit{reactive} policies.}

In \cref{sec:Policies Trained on Markovian Expert Demonstrations Exhibit Optimal Execution Horizon 1}, we argue that long execution horizons primarily mitigate the planning ambiguity that arises when short-context policies imitate non-Markovian expert behavior. We thus hypothesize that increasing context length should disproportionately improve policy performance at short execution horizons and shift the optimal horizon toward $T_{\text{exec}}^*=1$. Ablations across all 4 simulation benchmarks and 2 additional real-world tasks consistently support this hypothesis, with sufficiently long contexts making closed-loop execution optimal or near-optimal. At high data scales, long-context, short-execution-horizon policies even outperform short-context policies that rely on longer open-loop execution.

\subsection{Methodology}

Contemporary techniques for training policies that condition on a large number of raw image observations are still evolving. In this section, we propose new heuristics related to visual encoding to improve long-context performance.

For \texttt{FurnitureSimOneLeg}, \texttt{GearInsertion}, and the two real-world tasks, following \cite{agarwal2026trainingevaluatingdiffusionpolicies}, we use a U-Net Diffusion Policy architecture with cross-attention conditioning. We also apply a new method we call the ``double encoder", where the policy maintains separate image encoders (with separate parameters) for short-range and long-range observations. The most recent \texttt{short\_range} images are encoded by both encoders, and all earlier frames are encoded only by the long-range encoder; the U-Net conditions on all features via cross-attention. Intuitively, the double encoder allows the policy to learn separate representations for immediate control versus long-range planning and inferring expert hidden state. To prevent the policy from relying exclusively on the short-range pathway and ignoring long-range features, we also apply a dropout on the short-range encoder output, replacing it with a learned ``null" feature vector with some probability during training. In practice, we use \texttt{short\_range = 2} and a dropout rate of 0.3 for all policies. See \cref{sec:appendix_long_context_training_method} for more details. Although the double encoder is a heuristic architectural choice, we find that it stabilizes the performance of the long-context policies, enables effective recall of semantic information even with very long context lengths, and gives more predictable performance results (see \cref{sec:appendix_long_context_training_analysis}).

For \texttt{PushT-D}, following \cite{torne2025learninglongcontextdiffusionpolicies, agarwal2026trainingevaluatingdiffusionpolicies}, we use the baseline FiLM-conditioned U-Net with frozen observation encoder initialized from a trained short-context ($T_o=2$) policy. For \texttt{Kitchen}, we use just the baseline FiLM-conditioned U-Net (fully fine-tuned for each context length). These simpler architectures proved sufficient for these tasks and show that our results generalize across architectures. As presented in \cite{agarwal2026trainingevaluatingdiffusionpolicies}, naive context length scaling can be sufficient for U-Net architectures in moderate-to-high data regimes. 

We otherwise use the same training and evaluation methodologies as described in \cref{sec:Policies Trained on Markovian Expert Demonstrations Exhibit Optimal Execution Horizon 1}.

\subsection{Results and Discussion}

\begin{figure}[h!]
    \centering
    \includegraphics[width=0.675\linewidth]{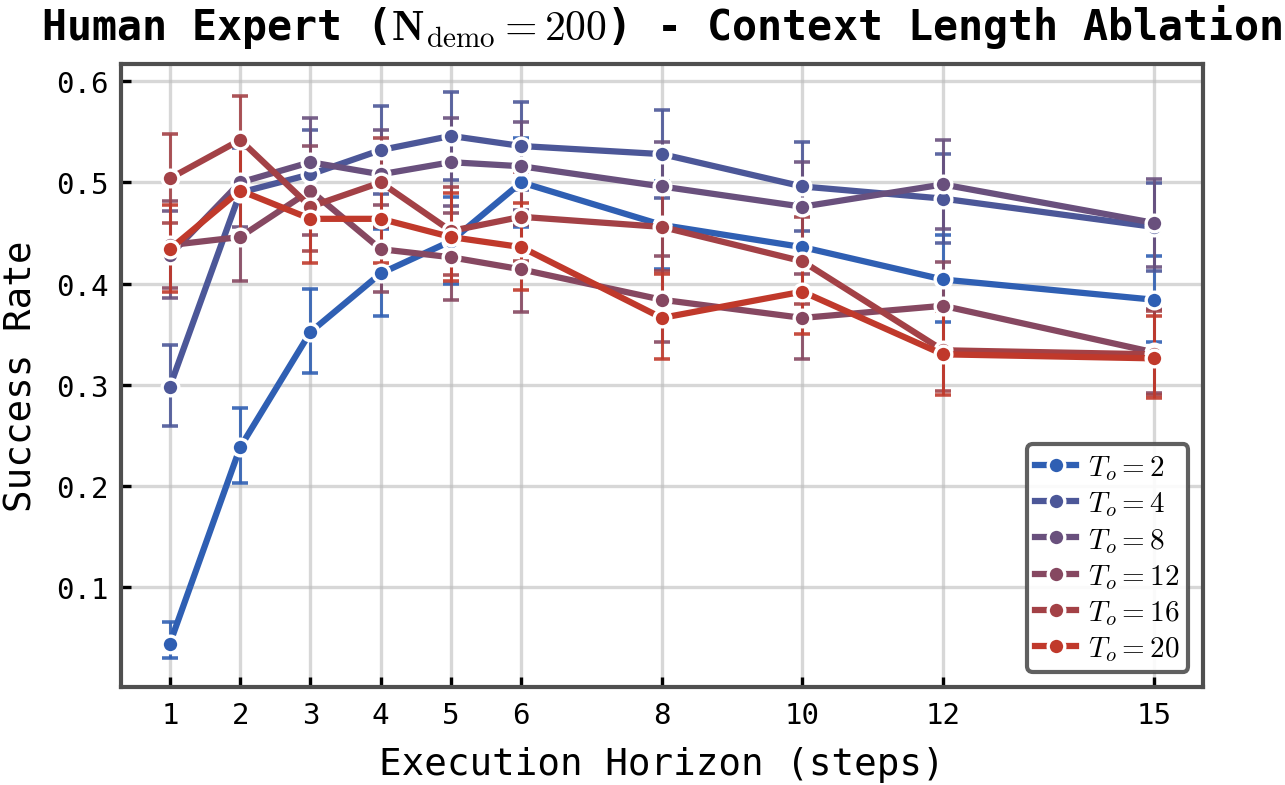}
    \caption{\textbf{Success-horizon curves of policies trained across a range of context lengths for \texttt{FurnitureSimOneLeg}, on 200 human demonstrations}. As context length increases, peak performance shifts toward shorter execution horizons, with the longest-context policies performing best at execution horizons of 1 or 2.} 
    \label{fig:context_length_ablation_human_expert_furnituresim}
\end{figure}

On \texttt{FurnitureSimOneLeg}, we first train Diffusion Policies with context lengths $T_o \in \{2, 4, 8, 12, 16, 20 \}$ on a dataset of 200 human demonstrations. The resulting success-horizon curves are shown in \cref{fig:context_length_ablation_human_expert_furnituresim}. With the standard $T_o=2$, the policy exhibits the typical inverted U-shaped success-horizon curve with intermediate execution horizons ($T_{\text{exec}}=6$) performing best. As $T_o$ increases, the optimal execution horizon $T_{\text{exec}}^*$ shifts left. At $T_o=$ 16 and 20, the curve becomes nearly monotonically decreasing with optimal execution horizon $T_{\text{exec}}^*=1$ or $2$, indicating that, at long context lengths, the highest-performing policies do not rely on long execution horizons and are the most reactive.

\begin{figure}[h!]
    \centering
    \includegraphics[width=1\linewidth]{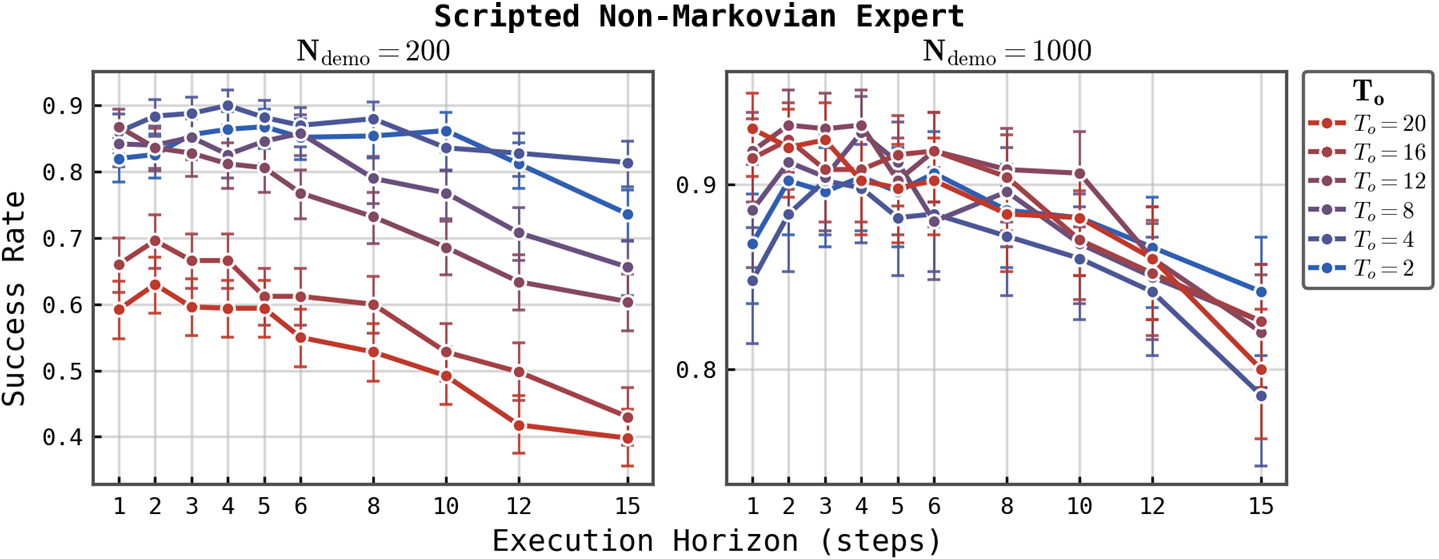}
    \caption{\textbf{Success-horizon curves of policies trained across a range of context lengths for \texttt{FurnitureSimOneLeg}, on 200 and 1000 non-Markovian scripted expert demonstrations.} We find increasing dataset size disproportionately improves performance of long-context policies at low execution horizons. With 1000 demonstrations, the highest performing policies use long context length $T_o$ of 12 -- 20 and short execution horizon $T_{\text{exec}}<4$ and achieve 93.2\% success, outperforming the best short-context ($T_o=2$) policy at $T_{\text{exec}}=6$ with 90.6\% success.}
    \label{fig:context_length_ablation_data_comparison}
\end{figure}

Next, because increasing context length substantially expands the policy's input space, increasing risk of encountering out-of-distribution inputs during rollout, we hypothesized that increasing dataset size would continue to improve performance at low execution horizons \cite{ agarwal2026trainingevaluatingdiffusionpolicies}. Therefore, we repeat the experiment with 200 and 1000 demonstrations generated by our non-Markovian scripted expert and compare results in \cref{fig:context_length_ablation_data_comparison}.\footnote{We note that we should expect a context length of 30 frames to be required to nearly reproduce the expert, as the injected history-dependent behaviors can span up to 30 frames (see \cref{sec:markovian_experts_experimental_setup}). Some latent variables may remain fundamentally unobservable, although longer context provides more information from which they may be inferred.} We observe the same qualitative trend: increasing context length disproportionately improves performance at lower execution horizons, shifting the optimal execution horizon $T_{\text{exec}}^*$ left. Further, comparing the success-horizon curves at 200 and 1000 demonstrations shows that increasing dataset size directly amplifies this trend. At 200 demonstrations, long-context policies with $T_{\text{exec}}<4$ still underperform, but at 1000 demonstrations, the highest-performing policies use long context lengths with highly reactive execution and significantly outperform the best $T_o=2$ policy.

Evaluating long-context policies on the remaining three simulation benchmarks: \texttt{PushT-D}, \texttt{GearInsertion}, and \texttt{Kitchen}, we find the same pattern holds: across all tasks, as context length is scaled high enough for the given task, the success-horizon curve becomes nearly monotonically decreasing and the most reactive policy is optimal or near-optimal. These results are consistent with our Expert Markovianity hypothesis (\cref{sec:Expert Markovianity Dominates the success-horizon curve}): longer history makes more of the expert’s latent, history-dependent state inferable to the policy, reducing the need to preserve action commitment through long execution horizons.

\begin{figure}[h!]
    \centering
    \includegraphics[width=1\linewidth]{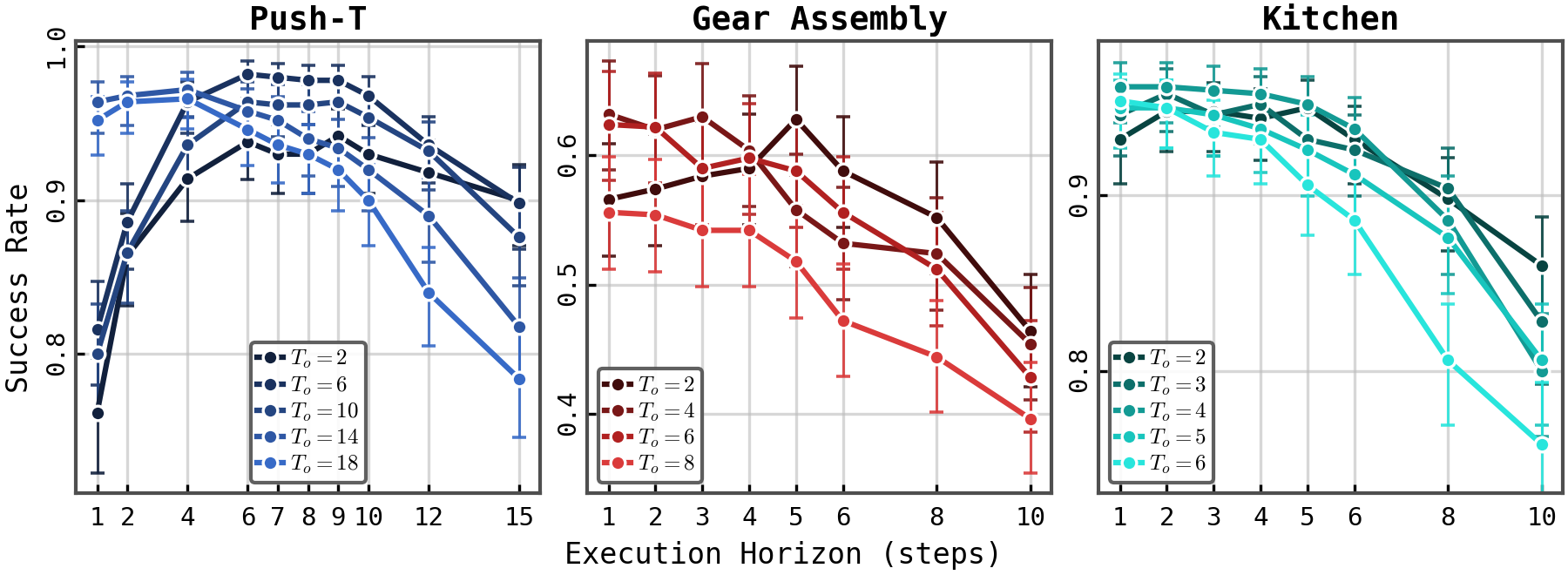}
    \caption{\textbf{Success-horizon curves of policies trained across a range of context lengths for \texttt{PushT-D}, \texttt{GearInsertion}, and \texttt{Kitchen} on human demonstrations}. Dataset sizes are 160 demonstrations for \texttt{PushT-D}, 200 for \texttt{GearInsertion}, and 581 for \texttt{Kitchen}. With sufficiently long context length, $T_{\text{exec}}=1$ becomes optimal or near-optimal in all tasks.}
    \label{fig:context_length_ablation_3_tasks}
\end{figure}

\begin{figure*}[h!]
    \centering
    \includegraphics[height=5.475cm]{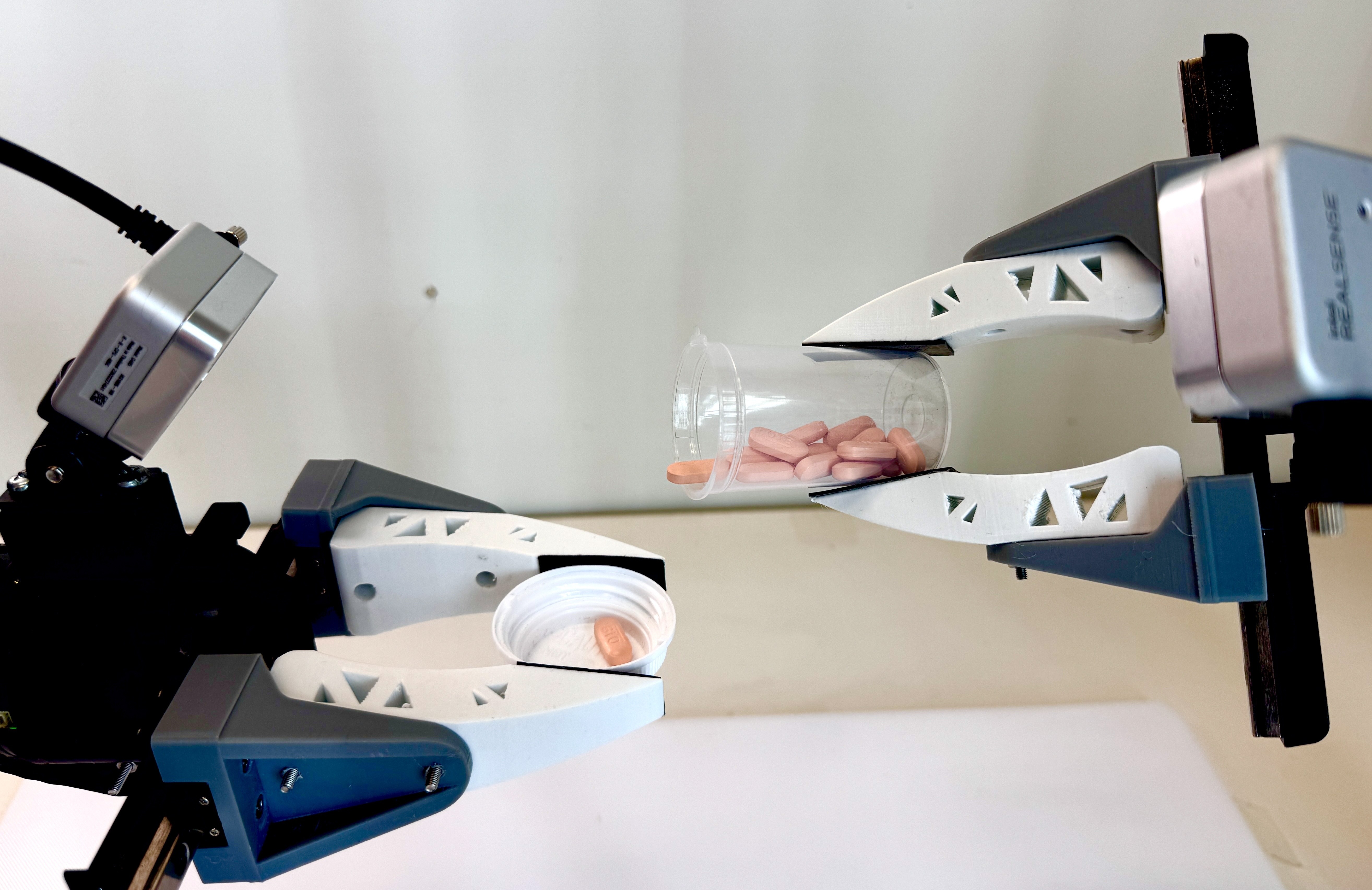}%
    \hspace{0.2mm}%
    \includegraphics[height=5.475cm]{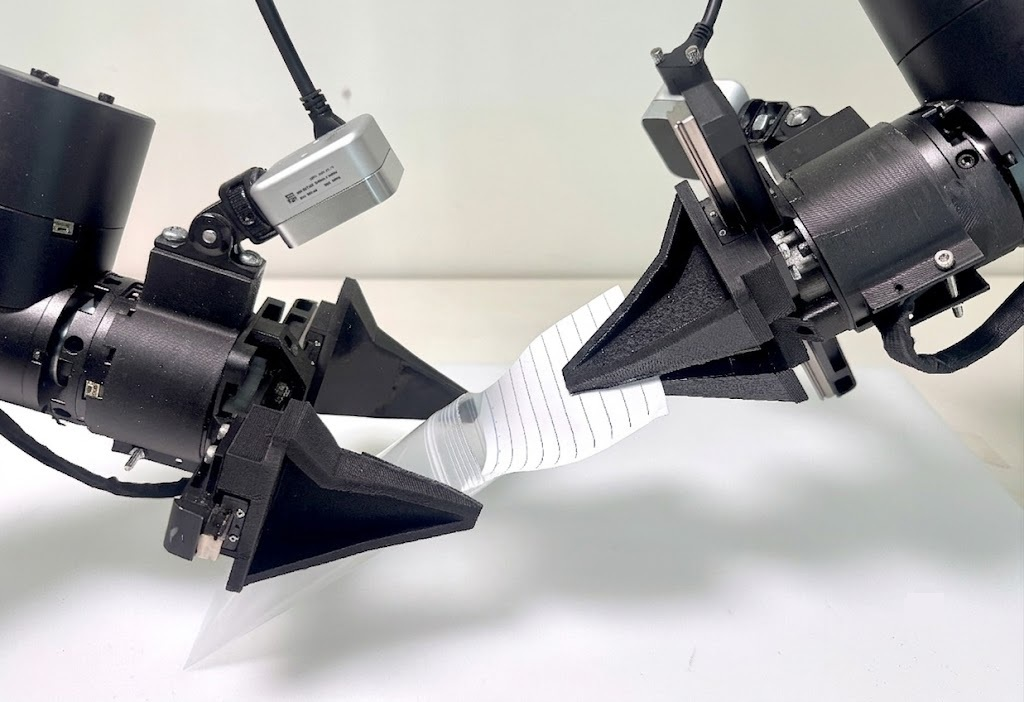}%
    \caption{\textbf{Real-world \texttt{SinglePillDispense} (Left) and \texttt{SlipIntoBaggie} (Right) tasks.} In \texttt{SinglePillDispense}, a successful rollout dumps a single pill out of the bottle onto the cap and then rights the bottle. In \texttt{SlipIntoBaggie}, a successful rollout fully inserts the paper slip into the baggie without crumpling the paper.}
    \label{fig:pill_and_baggie_tasks}
\end{figure*}

Lastly, we evaluate long-context policies on two real-world, highly-dexterous bimanual tasks: \texttt{SinglePillDispense} and \texttt{SlipIntoBaggie}, illustrated in \cref{fig:pill_and_baggie_tasks}.

Both of these tasks require careful, reactive dexterity  --- \texttt{SinglePillDispense} requires gentle tilting and intentional shaking to shift the pills toward the edge of the bottle without dumping out more than 1; \texttt{SlipIntoBaggie} requires careful shimmying to fit the paper slip into the baggie, which is barely larger than the paper. We train \texttt{SinglePillDispense} policies on just 1.34 hours of teleop data (516 demonstrations) and \texttt{SlipIntoBaggie} policies on 5.24 hours of data (540 demonstrations). We assess $T_o \in \{2, 8, 12\}$ and $T_{\text{exec}} \in \{2,4,8\}$, and present results in \cref{tab:hardware_context_length_ablation}. See \cref{sec:appendix_a_hardware_experiment_details} for additional details.

As in the simulation experiments, increasing context length from $T_o=2$ to $T_o=8$ shifts the optimal execution horizon from $T_{\mathrm{exec}}^*=8$ to $T_{\mathrm{exec}}^*=2$ while improving absolute performance. Reactivity appears to be highly beneficial in both tasks --- in \texttt{SinglePillDispense}, to stop dumping as soon as one pill is dispensed, and in \texttt{SlipIntoBaggie}, to react to unexpected paper or baggie deformities. However, reactive, short-execution-horizon policies do not perform well without extending context length to retain the behavioral coherence that long execution horizons would otherwise provide. These results show that our findings generalize under real-world visual distribution shifts, variable dynamics, and inference latency: long context lengths enable short execution horizons and outperform short-context, long-execution-horizon baselines.

\begin{table}[h!]
    \centering
    \caption{\texttt{SinglePillDispense} and \texttt{SlipIntoBaggie} success rates over 20 trials, with 95\% confidence intervals.}
    \label{tab:hardware_context_length_ablation}
    \setlength{\tabcolsep}{5pt}
    \begin{tabular}{@{}l ccc ccc@{}}
    \toprule
    & \multicolumn{3}{c}{\texttt{SinglePillDispense}}
    & \multicolumn{3}{c}{\texttt{SlipIntoBaggie}} \\
    \cmidrule(lr){2-4} \cmidrule(l){5-7}
    & $T_{\mathrm{exec}}=2$
    & $T_{\mathrm{exec}}=4$
    & $T_{\mathrm{exec}}=8$
    & $T_{\mathrm{exec}}=2$
    & $T_{\mathrm{exec}}=4$
    & $T_{\mathrm{exec}}=8$ \\
    \midrule
    $T_o=2$
    & 50\% {\scriptsize [29.9, 70.1]}
    & 60\% {\scriptsize [38.7, 78.1]}
    & 75\% {\scriptsize [53.1, 88.8]}
    & 60\% {\scriptsize [38.7, 78.1]}
    & 40\% {\scriptsize [21.9, 61.3]}
    & 80\% {\scriptsize [58.4, 91.9]} \\
    $T_o=8$
    & \extrabold{90\% {\scriptsize [69.9, 97.2]}}
    & 80\% {\scriptsize [58.4, 91.9]}
    & 60\% {\scriptsize [38.7, 78.1]}
    & \extrabold{85\% {\scriptsize [64.0, 94.8]}}
    & 70\% {\scriptsize [48.1, 85.5]}
    & 55\% {\scriptsize [34.2, 74.2]} \\
    $T_o=12$
    & 75\% {\scriptsize [53.1, 88.8]}
    & 70\% {\scriptsize [48.1, 85.5]}
    & 60\% {\scriptsize [38.7, 78.1]}
    & 75\% {\scriptsize [53.1, 88.8]}
    & 65\% {\scriptsize [43.3, 81.9]}
    & 65\% {\scriptsize [43.3, 81.9]} \\
    \bottomrule
    \end{tabular}
\end{table}

Finally, we note that performance declines at $T_o=12$ in both \texttt{SinglePillDispense} and \texttt{SlipIntoBaggie}, reflecting a broader pattern across our experiments: once the benefits of additional history saturate, the increased sample complexity of longer context lengths degrades performance. At the data scales used in our experiments, this degradation usually occurs only beyond the context length needed for a low-execution-horizon, reactive policy to achieve the highest absolute performance, and higher data scales extend the range of beneficial context lengths. However, determining the optimal tradeoff, and how it scales with data, remains future work.

\begin{takeawaybox}
\textbf{Key Takeaway.}

Increasing context length shifts peak performance toward shorter execution horizons, with sufficiently long context lengths reaching optimal $T^*_{\text{exec}}$ of 1 -- 2. Long-context, reactive control also scales favorably with data, outperforming short-context, long-execution-horizon baselines at sufficiently high data scales.
\end{takeawaybox}

Taken together, \textbf{these results establish long-context, reactive control as a viable and higher-performing alternative to long, open-loop execution} in the moderate-to-high-data regime.

\section{Discussion and Conclusion}
\label{sec:Discussion}
In this section, we conclude with additional discussion and directions for future research in robotic imitation learning.

Firstly, \textbf{we emphasize that data scale is an important experimental variable}. With the exception of the dataset-size ablation in \cref{fig:data_ablation} (and the \texttt{Push-T} context length experiments in \cref{fig:context_length_ablation_3_tasks}), all of our experiments use at least 200 demonstrations. We chose to study this moderate-to-high-data regime because it more closely reflects deployment-oriented robotics, where training on large datasets is increasingly necessary and common. Our ablation illustrates why this matters: with only 100 demonstrations, even a policy imitating a fully observable, deterministic, Markovian expert requires non-trivial execution horizon, whereas this preference disappears as dataset size increases. Because datasets of 50 -- 100 demonstrations remain common in robotics research, \textbf{we encourage evaluating results across dataset sizes to ensure results hold at scale.}

Secondly, we believe that the prevailing approach to policy training --- combining short context lengths with long open-loop execution horizons --- is poorly aligned with both human behavior and the desired operation regime of responsive robotic systems. While long execution horizons were previously recommended to avoid causal confusion risks of history-conditioning \cite{zhao2023learningfinegrainedbimanualmanipulation}, advances in data scale, compute, and policy architectures have since changed this tradeoff. Recent work shows that even naively extending context length is substantially less brittle than previously believed if appropriate architectures are selected \cite{agarwal2026trainingevaluatingdiffusionpolicies}; and there are many other avenues to pursue long-context learning \cite{torne2025learninglongcontextdiffusionpolicies, mark2026bpplongcontextrobotimitation, agarwal2026trainingevaluatingdiffusionpolicies, sridhar2025memerscalingmemoryrobot, torne2026memmultiscaleembodiedmemory}. \textbf{We therefore encourage practitioners to train policies with longer contexts --- even modestly longer with 4 -- 8 frames --- and evaluate correspondingly shorter execution horizons.}

Finally, we observe that the current literature in long-context learning has focused on ``global", task-relevant observability and memory; things like selecting the correct manipuland, counting repetitions, or recalling the location of hidden objects (i.e. in the ``shell game" \cite{guo2026chameleoncontrolindexedprospective,rhoda2026dva}). Accordingly, many methods represent observations through compact language-like tokens or summaries while retaining only the most recent visual frame as direct observation input. Our results highlight a more subtle role of context that such summaries may not fully capture: by resolving hidden-state aliasing and enabling shorter execution horizons, \textbf{raw, visual context is critical for reactive, coherent, ``local" control}. Yet, exploration of such context has been limited to only a few works \cite{agarwal2026trainingevaluatingdiffusionpolicies, torne2025learninglongcontextdiffusionpolicies}. While semantic memory is important for long-horizon manipulation, we believe context-encoding methods should also be evaluated on dexterous tasks under short execution horizons. A method of encoding context that effectively recalls locations of hidden objects but does not enable effective manipulation with an execution horizon of 1, we argue, is missing an important dimension of temporal understanding.

\clearpage
\acknowledgments{
We thank Leslie Kaelbling and Anthony Simeonov for thoughtful discussions and feedback and John Marangola for suggestions on our teleop interface. Hardware experiments for this project were supported by Mundane Systems Inc.

This material is based upon work supported by Amazon.com Services LLC PO \#2D-19307345, the National Science Foundation (NSF) Engineering Research Center for Human AugmentatioN via Dexterity (HAND) (Grant No. 2330040), and Toyota Research Institute PO \#003859. Any opinions, findings, and conclusions or recommendations expressed in this material are those of the author(s) only. 

We thank The Infrastructure Group at MIT CSAIL for maintaining the computing resources primarily used for this work. We also acknowledge the MIT SuperCloud and Lincoln Laboratory Supercomputing Center for providing HPC, database, and consultation resources \cite{Reuther_2018_supercloud}.
}

\bibliography{ref}  %

\appendix
\section{Experimental Setup}
\label{sec:appendix_a_experimental_setup}

\subsection{Benchmark Simulation Task Details}
\label{sec:appendix_a_benchmark_task_details}

For \texttt{FurnitureSimOneLeg} \cite{heo2023furniturebenchreproduciblerealworldbenchmark}, the goal of the task is to push the table top into the obstacle corner, pick up the table leg, then screw the table leg into the table top. The poses of the table top and table leg are randomized at the start of the episode using the ``low" randomization setting provided by FurnitureSim \cite{heo2023furniturebenchreproduciblerealworldbenchmark}. The scene has two cameras, a static front-facing camera and a wrist-mounted camera. Proprioceptive state and action space and policy predicted-action frequencies for all tasks are summarized in \cref{tab:benchmark_observation_action_spaces,tab:benchmark_policy_frequencies}. Our expert comparison experiments (\cref{fig:human_vs_markovian_vs_non_markovian_furnituresim}) use datasets of 200 demonstrations from each expert type. We use a SpaceMouse and adapt code from \cite{ankile2024juicerdataefficientimitationlearning} to collect human demonstrations.

In \texttt{Push-T}, the goal of the task is to push the T-shaped slider from a randomly initialized pose to a fixed goal pose. The robot is constrained to motion within the XY plane with a fixed end-effector Z coordinate and uses a cylindrical tool that allows only approximately point contacts with the T. We created a planar variant of the ManiSkill \cite{mu2021maniskillgeneralizablemanipulationskill} \texttt{Push-T} environment, use the ManiSkill RL training pipeline to train our Markovian expert (see \cref{sec:appendix_a_markovian_experts} for more detail), and evaluate performance between Markovian and human experts (\cref{fig:human_vs_markovian_3_tasks}) in ManiSkill. However, for our \texttt{Push-T} long-context ablation (\cref{fig:context_length_ablation_3_tasks}), we use the \texttt{Push-T} environment in Drake \cite{tedrake2019drake} (as well as the open source dataset) provided by \cite{wei2025empiricalanalysissimandrealcotraining}. The ManiSkill and Drake task variants are identical beyond simulation parameters except that the Drake implementation adds an additional requirement that the robot must return to its start position after pushing the T for success. The reason for the difference in simulators is that Drake provides a richer hydroelastic contact model \cite{masterjohn2022velocitylevelapproximationpressure} that makes it well-suited for planar pushing, but training an RL expert is much more amenable in ManiSkill due to its GPU parallelism and simulation speed. Both scenes use two cameras, a static front-facing camera and a wrist-mounted camera. Our human vs Markovian expert comparison (\cref{fig:human_vs_markovian_3_tasks}) uses datasets of 300 demonstrations for each while our context length ablation (\cref{fig:context_length_ablation_3_tasks}) uses 160 demonstrations.

For \texttt{GearInsertion}, the goal of the task is to pick up a gear initialized at a random pose and insert it onto a gear carrier, meshing it with the other gears. We use an adaptation of the ``Factory Gear Mesh" task in IsaacLab \cite{nvidia2025isaaclabgpuacceleratedsimulation, narang2022factoryfastcontactrobotic}. The scene has three cameras, a static front-facing camera, a static rear-left, upward-angled camera, and a wrist-mounted camera. The rear-left camera is added to improve visibility of the shaft and gear bore during insertion. All \texttt{GearInsertion} experiments (\cref{fig:human_vs_markovian_3_tasks,fig:context_length_ablation_3_tasks}) use datasets of 200 demonstrations. The ``human" expert dataset consists of 100 demonstrations collected using a SpaceMouse and 100 additional demonstrations synthetically generated using MimicGen \cite{mandlekar2023mimicgendatagenerationscalable}. The Markovian expert dataset consists of 200 successful rollouts from our Markovian scripted expert as described in \cref{sec:appendix_a_markovian_experts}.

For \texttt{Kitchen}, the goal of the task is to complete any 4 subtasks, among 7 total: opening the microwave, moving the kettle to the far burner, turning the top or bottom knob, flipping the light switch, opening the left cupboard door, and sliding open the right cupboard door. All subtasks begin in their incomplete states. Success is determined when any 4 are completed. We use the environment from \cite{gupta2019relaypolicylearningsolving} (with 2 added cameras --- a scene camera and wrist camera) as well as the publicly released 581 human demonstrations (which we re-rendered with camera observations) combined with 581 Markovian scripted expert demonstrations that we generated for all experiments (\cref{fig:human_vs_markovian_3_tasks,fig:context_length_ablation_3_tasks}).

\begin{table}[h!]
    \centering
    \caption{Benchmark Tasks -- Proprioceptive Inputs and Action Spaces}
    \label{tab:benchmark_observation_action_spaces}
    \begin{tabular}{@{}p{0.22\columnwidth}p{0.37\columnwidth}p{0.34\columnwidth}@{}}
        \toprule
        Task & Proprioceptive State & Action Space \\
        \midrule
        \texttt{FurnitureSimOneLeg}
        & 16D: end-effector pose, velocity, and gripper width
        & 10D: delta end-effector command (3D translation, 6D rotation, 1D gripper) \\

        \texttt{Push-T}
        & 3D: planar end-effector position and orientation
        & 2D: delta end-effector position in the XY plane \\

        \texttt{GearInsertion}
        & 9D: 3D end-effector position, 4D orientation quaternion, 2D gripper state
        & 7D: delta end-effector pose and gripper command \\

        \texttt{Kitchen}
        & 9D: seven arm-joint positions and two gripper-finger positions
        & 9D: velocity commands for the seven arm joints and two gripper joints \\
        \bottomrule
    \end{tabular}
\end{table}

\begin{table}[h!]
    \centering
    \caption{Benchmark Tasks -- Policy Action Prediction Frequencies}
    \label{tab:benchmark_policy_frequencies}
    \begin{tabular}{@{}p{0.3\columnwidth}p{0.24\columnwidth}@{}}
        \toprule
        Task & Predicted-Action Frequency \\
        \midrule
        \texttt{FurnitureSimOneLeg} & 10 Hz \\
        \texttt{Push-T}             & 10 Hz \\
        \texttt{GearInsertion}      & 30 Hz \\
        \texttt{Kitchen}            & 12.5 Hz \\
        \bottomrule
    \end{tabular}
\end{table}

\subsection{Markovian Experts for \texttt{Push-T}, \texttt{GearInsertion}, and \texttt{Kitchen}}
\label{sec:appendix_a_markovian_experts}

For \texttt{Push-T}, we train a Markovian Reinforcement Learning (RL) policy and evaluate it deterministically to generate Markovian expert data. The RL policy is parameterized as a Gaussian policy with an MLP and is trained from scratch using PPO \cite{schulman2017proximalpolicyoptimizationalgorithms}.

For \texttt{GearInsertion}, like for \texttt{FurnitureSimOneLeg}, we develop a Markovian scripted expert that uses a finite-state-machine (FSM) that determines its action using only the current environment state. For \texttt{GearInsertion}, in order to introduce some diversity during insertion (to prevent perfect insertions during all demonstrations, which yields a very thin training data manifold that the learner would easily step off of during rollout), we inject a small amount of Gaussian noise into both the executed and recorded actions proposed by the expert.

For \texttt{Kitchen}, we develop a deterministic Markovian scripted FSM-based expert for each of the 7 individual subtasks. During rollout, to complete sequences of 4 subtasks, we chain 4 of the subtask policies together with return-to-home primitives connecting them.

Because \texttt{Kitchen} permits completing any sequence of 4 of the 7 subtasks, the environment state alone does not uniquely specify which subtask the expert should execute next. To generate diverse Markovian expert demonstrations covering different subtask permutations, we sample a four-subtask sequence before each rollout and record it as an additional conditioning variable. Both the scripted expert and the Diffusion Policy have access to this sequence, making the expert's current and future subtask intent observable to the learner. Without this conditioning variable, randomly selecting a sequence would introduce latent expert state: identical environment states could produce different actions depending on a previously selected but unobserved subtask ordering. Note that this subtask sequence label is used only for the Markovian scripted-expert dataset and is not provided during human teleoperation. 

To match \texttt{Kitchen}'s human teleop data collection environment (and the publicly released demonstration data), we also add random noise to the recorded observations during scripted expert rollout.

For all tasks, we roll out the Markovian expert and retain only successful episodes in the dataset. Conditioning on success may render the resulting induced expert non-Markovian; however, because our Markovian experts achieve high success rates (above 90\%), we expect this effect to be limited.

\subsection{Hardware Experiment Details}
\label{sec:appendix_a_hardware_experiment_details}

\paragraph{\texttt{SinglePillDispense}.}
The robot begins holding the pill bottle cap in its left gripper and pill bottle in its right gripper and must dispense exactly one pill into the cap within 30 s. Dispensing more or less than 1 pill constitutes failure. The time limit prevents incidental pill motion caused by prolonged, incoherent robot jitter from being counted as success. Each pill weighs approximately 0.6 g., and the bottle contains up to 24 pills and measures 1.7 in. diameter by 2.8 in. length.

\paragraph{\texttt{SlipIntoBaggie}.}
The robot must pick up a plastic baggie from the table and a paper slip from an elevated surface, then insert the slip into the baggie within 2 min. Success requires the slip to remain uncreased; an episode is terminated and marked as failure if the slip is creased or actively being creased. The slip measures $83.7 \times 73.8$ mm ($\pm 2$ mm), and the baggie measures $127 \times 76.2$ mm.

\paragraph{Hardware and control implementation.}
Both tasks use two ARX-5 follower arms with custom grippers, teleoperated using two ARX-5 leader arms. Observations comprise RGB images from two wrist-mounted Intel RealSense D405 cameras and one overhead Intel RealSense D435 camera. We use the same training procedure as for the simulation tasks and train Diffusion Policies with delta end-effector-pose actions and 10 Hz actions and observations. A 125 Hz linear interpolator generates intermediate target poses between the policy-predicted 10 Hz actions, which are converted to target joint positions by an inverse-kinematics solver and executed by a low-level joint-position PD controller. During evaluation, we apply two very simple but effective latency-control mechanisms. Firstly, we cache observation embeddings for future inference calls to re-use, preventing redundant forward passes of the observation encoder. Secondly, to avoid discontinuities between action chunks caused by inference latency, as discussed in \cref{sec:related_works}, we linearly interpolate from the final action of the previous chunk to the action 180 ms into the newly predicted chunk before executing the remainder of the new chunk (the value of 180 ms was tuned empirically).

\paragraph{Evaluation procedure.} During training, we save the 5 checkpoints with lowest DDIM MSE loss on the validation set. We screen each checkpoint using a small set of preliminary hardware rollouts, considering both task performance and qualitative stability. We then evaluate the chosen checkpoint on an independent set of 20 additional trials. We note that checkpoint selection was particularly important at long context lengths -- we discuss this further in \cref{sec:appendix_checkpoint_selection}.

\paragraph{Note on lighting variation.}
We note that color-jitter augmentations were \textit{unintentionally} omitted during training. To limit lighting variation, all evaluations were conducted within the same five-hour window each day. Nevertheless, demonstrations were collected under varying lighting conditions, potentially adding extra visual variation into the dataset and reducing the effectiveness of some data samples. We expect that re-training with stronger lighting invariance would improve performance, particularly for longer-context policies that are already more subject to out-of-distribution inputs.

\section{Training and Evaluation Methods}
\label{sec:appendix_a_training_and_evaluation_methods}

\subsection{Diffusion Policy}
\label{sec:appendix_difffusion_policy}

Diffusion Policy \cite{chi2024diffusionpolicyvisuomotorpolicy}  represents $\pi_\theta(A_t \mid O_t)$ as a conditional denoising diffusion model over action chunks. We parameterize this distribution with a neural network $\epsilon_\theta$ comprising image encoders that map the observation history $O_{t}$ to observation features and a conditional U-Net that conditions on these observation features and iteratively denoises the action chunk.

During training, a clean expert chunk $A_t$ is corrupted as $$A_t^k = \alpha_k A_t + \sigma_k \epsilon, \qquad \epsilon \sim \mathcal N(0,I),$$

where $k$ indexes the diffusion noise level. The policy learns a family of denoisers, one for each noise level, typically by predicting the added noise:

$$\min_\theta \mathbb{E}_{(O_t,A_t)\sim\mathcal D,;k,;\epsilon} \left[ \left| \epsilon - \epsilon_\theta(A_t^k, k, O_t) \right|^2 \right].$$

At inference time, the policy initializes $A_t^K \sim \mathcal N(0,I)$ and iteratively denoises it to predict action chunk $\hat A_t$. The robot then executes the first $T_{\text{exec}}$ actions of $\hat A_t$ open-loop before re-querying the policy.

\subsection{Training and Evaluation Parameters}

In all simulation experiments, we use ResNet-18~\cite{he2015deepresiduallearningimage} as the image encoder and fine-tune its weights during training. For \texttt{FurnitureSimOneLeg}, \texttt{GearInsertion}, and \texttt{Kitchen}, we initialize the encoder with R3M-pretrained weights~\cite{nair2022r3muniversalvisualrepresentation}. We use R3M because the ablations in \cite{ankile2024juicerdataefficientimitationlearning} show that it outperforms ImageNet-1K, CLIP, and other initialization methods on FurnitureSim tasks. For \texttt{Push-T}, we instead use the ImageNet-1K-pretrained weights provided by robomimic~\cite{mandlekar2021matterslearningofflinehuman} and replace the encoder's BatchNorm layers with GroupNorm, following Diffusion Policy~\cite{chi2024diffusionpolicyvisuomotorpolicy}. For the hardware experiments, \texttt{SinglePillDispense} and \texttt{SlipIntoBaggie}, we use a larger, pre-trained DINOv3 \cite{simeoni2025dinov3} vision transformer and fine-tune during training with $\frac{1}{10}$ of the policy learning rate. We find superior performance with DINOv3 (compared to CLIP or ResNet-based image encoders) in high precision real-world tasks.

For inputs containing $C$ cameras and $T_o$ observation timesteps, we use $C$ separate image encoders, one for each camera, and share each encoder's weights across the corresponding $T_o$ timesteps for training efficiency. All remaining shared training and model parameters are reported in \cref{tab:shared_training_hyperparameters,tab:diffusion_model_hyperparameters}.

\begin{table}[h!]
    \centering
    \caption{Training Hyperparameters Shared for All Models}
    \label{tab:shared_training_hyperparameters}
    \begin{tabular}{@{}p{0.40\columnwidth}p{0.20\columnwidth}@{}}
        \toprule
        Parameter & Value \\
        \midrule
        Max LR                         & $10^{-4}$ \\
        LR Scheduler                   & Cosine \\
        Weight Decay                   & $10^{-6}$ \\
        Warmup Steps                   & 500 \\
        Batch Size                     & 64 \\
        Image Size Input               & $N_{\text{cameras}} \times 320 \times 240 \times 3$ \\
        Image Size Encoder             & $N_{\text{cameras}} \times 224 \times 224 \times 3$ \\
        Image Augmentations            & 95\% random crop \\
        Encoder Feature Projection Dim & 128 \\
        \bottomrule
    \end{tabular}
\end{table}

\begin{table}[h!]
    \centering
    \caption{Diffusion Model Hyperparameters}
    \label{tab:diffusion_model_hyperparameters}
    \begin{tabular}{@{}p{0.4\columnwidth}p{0.29\columnwidth}@{}}
        \toprule
        Parameter & Value \\
        \midrule
        U-Net Down dims                & $[256, 512, 1024]$ \\
        U-Net Parameters               & 69 million \\
        DDIM Inference Steps           & 10 \\
        \bottomrule
    \end{tabular}
\end{table}

For each dataset, we reserve 5\% of demonstrations for validation (except in \texttt{SinglePillDispense} and \texttt{SlipIntoBaggie}, where we reserve 2\%); thus, the effective training-set size is $0.95 \times$ ($0.98 \times$ for \texttt{SinglePillDispense} and \texttt{SlipIntoBaggie}) the stated number of demonstrations for each task. We intentionally train each policy well into the overfitting regime, where validation loss has increased substantially beyond its minimum, because Diffusion Policies have often been found to achieve their best rollout performance in this regime \cite{he2025demystifyingdiffusionpoliciesaction}.

During training, we save checkpoints every 5 epochs. We then select the 5 checkpoints with the lowest DDIM MSE loss on the validation set for closed-loop evaluation; we do not perform closed-loop evaluations during training. For each of the 5 chosen checkpoints and each execution horizon, we run 500 evaluation episodes, terminating each episode upon success or timeout. We set each task's timeout to approximately 120\% of the longest demonstration in its dataset. Error bars throughout the paper denote 95\% Wilson confidence intervals for the success probability of the reported (highest-performing) checkpoint for each task, architecture, and execution-horizon configuration; they characterize uncertainty from finite evaluation rollouts, but do not capture randomness from the training process. A more rigorous analysis with explicit hypothesis testing and correction for multiple policy comparisons as in \cite{trilbmteam2025carefulexaminationlargebehavior} is desirable but left to future work.

\section{Understanding the Benefits of Long Execution Horizons and Long Context}

\subsection{Why Long Execution Horizons Prevent Failure}
\label{sec:Why Long Execution Horizons Prevent Failure}

In this section, we provide informal conceptual explanations for understanding why long execution horizons prevent failure in the presence of non-Markovian demonstrations. For this section specifically, subscripts indicate time index, and superscripts are used to distinguish states distinct in task space.

\begin{figure}[h!]
    \centering
    \includegraphics[width=1\linewidth]{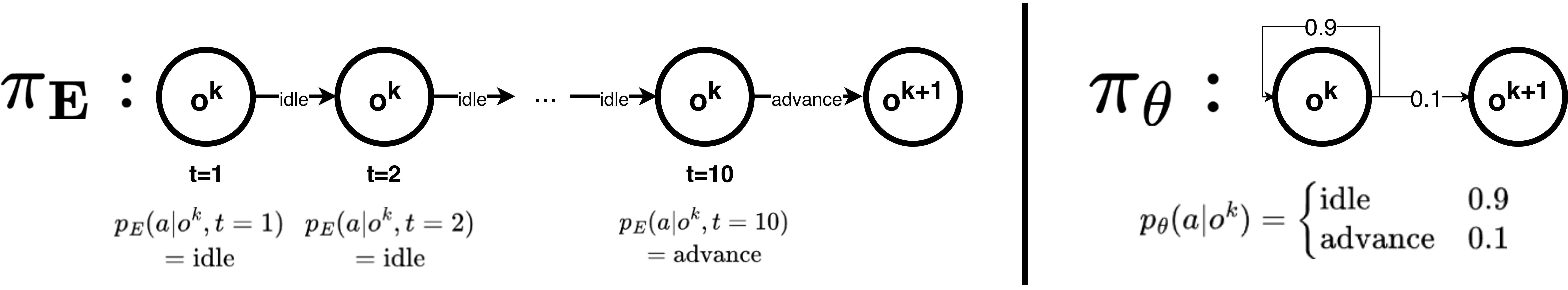}
    \caption{\textbf{Illustration of hidden-state aliasing in an idle demonstration.} While making high precision maneuvers in tasks like \texttt{Push-T}, humans often pause to align, observe, and think. Such pauses show up as a sequence of idle actions $o^k_1 \rightarrow o^k_2 \rightarrow \cdots \rightarrow o^k_{10}$ in the training data. In this example, the full-history conditioned expert $\pi_E$ advances after it observes the 10th idle state. The imitator $\pi_\theta$ only observes the current state $o_0$ and cannot infer idle count, so both idling and advancing appear valid. The imitator is therefore unlikely to reproduce the expert's exact sequence of 10 idle actions before advancing, and may idle for much longer than desired, resulting in task timeout.}
    \label{fig:action_chunking_theory_hidden_state_aliasing}
\end{figure}

Consider the phenomenon of \textit{hidden-state aliasing}: at a given state, conditioned on full history, the expert has a specific action distribution. However, the policy's short history may be consistent with multiple full histories in the training dataset that induce different expert action distributions. Even a well-trained short-context imitator may therefore sample an action from any one of these valid-appearing distributions and select an action inconsistent with the expert’s full-history-conditioned behavior. This can send the policy down an incorrect path or into a cycle. We illustrate this phenomenon with an example in \cref{fig:action_chunking_theory_hidden_state_aliasing}.

Long open-loop execution reduces the likelihood of incorrect action selection from hidden-state aliasing by reducing the probability of the policy re-planning at an aliased state. In particular, a policy with execution horizon $T_{\text{exec}}$ re-plans every $T_{\text{exec}}$ timesteps; assuming no correlation between phase with state, any aliased state has probability $\frac{1}{T_{\text{exec}}}$ of coinciding with a re-planning step. In the example in \cref{fig:action_chunking_theory_hidden_state_aliasing}, the policy can assuredly imitate the 10 idle actions if it executes the entire 10-step aliased sequence open-loop, without re-planning (for instance, if policy inference happens in the state right before $o^k$ and execution horizon is at least the length of the idle interval). If the policy re-plans at any intermediate step, it may wrongly predict to idle for too long or too short.

For mostly unimodal tasks like \texttt{FurnitureSimOneLeg} or \texttt{GearInsertion}, there may not necessarily be aliased states where some apparently valid action distributions lead to unrecoverable outcomes. In most cases, hidden-state aliasing is detrimental by inducing unintended \textit{cycles} --- any sequence of states $o_0^k \rightarrow o_1^{k+1} \rightarrow o_2^{k+2} \rightarrow \dots \rightarrow o_T^{k}$ that revisits a state --- that can lead to task timeout. In particular, by revisiting the aliased state $o^k$ multiple times, the policy may repeatedly select the locally-valid cycling action, keeping it in the cycle indefinitely. The simplest example of such a cycle would be an idle demonstration (i.e. \cref{fig:action_chunking_theory_hidden_state_aliasing}), where a policy that over-predicts idling may idle forever. We illustrate another example of cycling in \cref{fig:failure_cases}. Note that even when cycles do not exist in any single demonstration, they may appear in the cumulative dataset if the expert is non-Markovian, i.e. if one demonstration contains $o^k_t \rightarrow \dots \rightarrow o_{t+i_1}^{k+1}$, and another contains $o_{t}^{k+1} \rightarrow \dots \rightarrow o_{t+i_2}^k$. 

\begin{figure}[h!]
    \centering
    \includegraphics[width=0.5\linewidth]{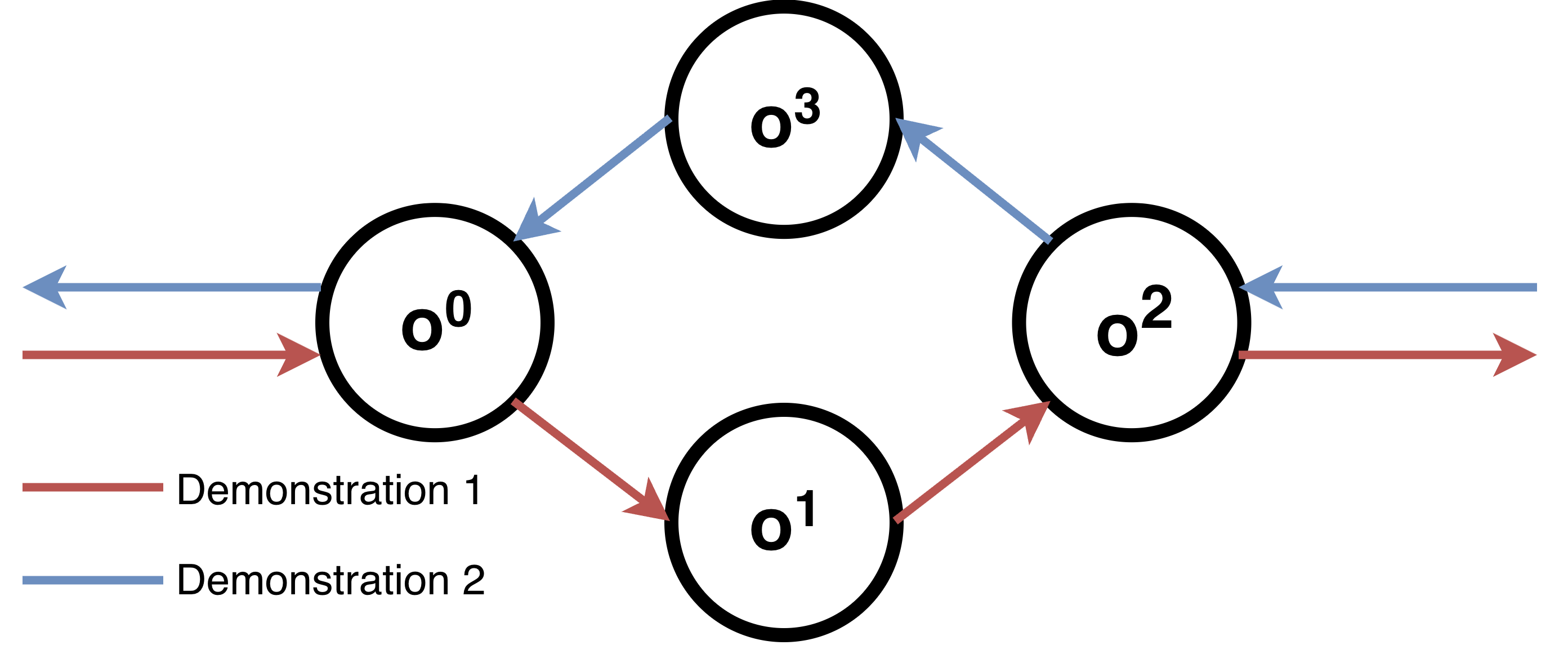}
    \caption{\textbf{Illustration of cycling.} Two demonstrations jointly induce a cycle in the aggregated dataset. When observing only $o^0$ or $o^2$, a short-context imitator cannot determine whether to continue or exit the cycle, since both actions appear valid in the training data. It may therefore repeatedly switch between the demonstrated trajectories and enter a cycle: $o^0 \rightarrow o^1 \rightarrow o^2 \rightarrow o^3 \rightarrow o^0$. Longer execution horizons reduce the chance of entering such a cycle: a chunk of actions originating at $o^0$ toward $o^1$ and executed open-loop with $T_{exec} \geq 3$ will cross beyond $o^2$, whereas re-planning at $o^2$ can increase the chance of returning to $o^3$ and entering a cycle.} 
    \label{fig:action_chunking_theory_looping}
\end{figure}

Crucially, using an execution horizon of $T_{\text{exec}}$ grants immunity from getting trapped in cycles of length $< T_{\text{exec}}$. Because the policy executes an entire $T_{\text{exec}}$-step action chunk open-loop, it traverses past the aliased state $o^k$ in one open-loop chunk and therefore cannot re-select the cyclic behavior. For cycles longer than $T_{\text{exec}}$, increasing $T_{\text{exec}}$ still reduces probability of cycling by simply decreasing the probability of re-planning on the aliased state to $\frac{1}{T_{\text{exec}}}$.

While the probability of repeating a cycle many times is theoretically very small, Diffusion Policies are known to overfit and learn high-Lipschitz functions rather than learning the true multimodal expert distribution \cite{pan2026adonoisingdispellingmyths}. In practice, when a policy repeats a cycle once, we find that it is likely to repeat it many times.

In summary: long open-loop execution can be thought of as a mechanism to reduce the probability of incorrect action selection from hidden-state aliasing and to prevent unbounded cycling.

\subsection{Why Long Context Lengths are a Principled Substitute for Long Execution Horizons}
\label{sec:appendix Why Long Context Lengths Outperform Long Execution Horizons}

\begin{figure}[H]
    \centering
    \includegraphics[width=0.6\linewidth]{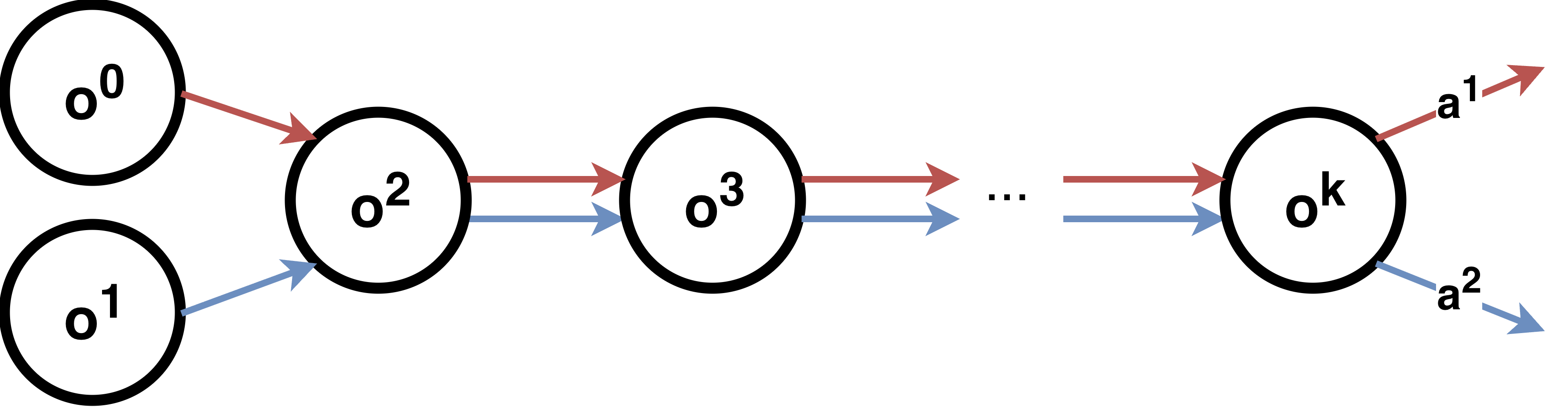}
    \caption{\textbf{Illustration of how context length disambiguates the expert action distribution.} The observations $o^2,\ldots,o^{k}$ form an aliased sequence: conditioned only on observations within this sequence, a short-context policy cannot identify the expert’s intended behavior (whether to imitate demonstration 1 or 2). A long execution horizon reduces the probability of re-planning within the sequence, but ambiguity remains whenever re-planning occurs there. By contrast, a sufficiently long context includes the preceding observation $o^0$, which disambiguates the correct expert action ($a^1$) distribution throughout the sequence, regardless of re-planning phase.
    }
    \label{fig:action_chunking_theory_context}
\end{figure}

In this section, we argue that long context lengths can serve the same role as long execution horizons when imitating non-Markovian demonstrations and, when they contain sufficient history, do so more directly. As established in \cref{sec:Why Long Execution Horizons Prevent Failure}, long execution horizons reduce the probability of hidden-state aliasing by reducing the frequency the policy re-plans at aliased states. Longer contexts instead preserve frequent re-planning while using preceding observations to disambiguate the expert's intended behavior at those aliased states. We illustrate this in \cref{fig:action_chunking_theory_context}.

The same distinction applies to cycling. Long execution horizons reduce the opportunity to re-select a cyclic action, while a context spanning the relevant history can reveal that the aliased state has already been visited and disambiguate the expert's exit behavior. 

In summary, long contexts address the same failure modes as long execution horizons, but do so in an arguably more principled way (that also maintains policy reactivity), by eliminating the underlying planning ambiguity rather than merely reducing how often the policy encounters it.

\section{HG-DAgger Implementation Details}
\label{sec:appendix_hg_dagger}

We provide additional implementation details for the HG-DAgger experiment in \cref{sec:Investigating the Role of Compounding Errors}. For each expert type, all execution horizons were initially evaluated with the same base policy, but HG-DAgger was thereafter run independently for each execution horizon. Each execution horizon-branch maintained its own aggregated dataset, checkpoints, failure rollouts, and correction trajectories; correction data were never shared across execution horizons.

In each round, we rolled out the best-performing checkpoint from the preceding round until obtaining 50 failed episodes. A human annotator labeled the \emph{gating timestep} of each failure, defined as the first timestep at which the policy deviated from a nominal trajectory in a manner that ultimately caused failure. For policies trained on non-Markovian expert data, the annotator also labeled the finite-state-machine state from which to initialize the scripted expert. The expert was then rolled out from the labeled state until success; episodes were discarded after 10 consecutive failed correction attempts. Provided at least 40 corrections succeeded, we added 40 to the corresponding branch's dataset, retrained its policy for 50,000 steps, and evaluated the resulting checkpoints before the next round. We chose to conduct 3 rounds as this was sufficient for convergence for the experiments in \cite{kelly2019hgdaggerinteractiveimitationlearning} and allows us to assess early trends in the success-horizon curve, when the effects of DAgger are greatest.

Because HG-DAgger must be run separately for each value of $T_{\text{exec}}$, we only evaluated 5 values of $T_{\text{exec}}$, instead of the $T_{\text{exec}} \in \{1,2,3,4,5,6,8,10,12,15\}$ sweep used in other experiments. We note that, with 3 rounds of HG-DAgger, 50 failed rollouts collected at each round, for 5 values of $T_{\text{exec}}$, for both the Markovian and non-Markovian expert: a total of 1500 episodes had to be manually annotated, by a single annotator for consistency.

\section{Exploratory Noise-Injection Experiment}
\label{sec:appendix_noise_injection}

Expanding on \cref{sec:Investigating the Role of Compounding Errors}, we evaluate one more intervention designed to mitigate compounding errors proposed by \citet{zhang2025actionchunkingexploratorydata}: exploratory noise injection; and we show it's impact on the success-horizon curve. 

Exploratory noise injection mitigates compounding errors by injecting noise during expert data collection to expand supervision around the expert distribution. We mimic the formulation in \cite{zhang2025actionchunkingexploratorydata}, where a fraction $\alpha$ of trajectories are collected cleanly, while the remaining trajectories execute perturbed expert actions $u_t=\pi^\star(x_t)+\sigma_u z_t$ but record the clean expert label $\pi^\star(x_t)$.\footnote{\label{fn:noise-injection-chunking-caveat} As noted by Zhang et al., this intervention has an important caveat in chunked prediction. For one-step behavior cloning, every perturbed state is paired with a corrective expert action. With multi-step prediction, however, later expert action labels are evaluated at states reached under noisy actions rather than under the preceding clean action labels in the target chunk. This state–action-sequence mismatch can make useful exploratory coverage appear as target uncertainty, particularly for long prediction horizons.} In our experiment, we treat $\sigma_u$ as a scalar, multiplied to state-dependent noise standard deviations, because different states require different levels of precision. We apply this technique during data collection using our non-Markovian scripted expert. While Zhang et al.'s theoretical analysis for noise injection require deterministic Markovian experts, we discuss caveats with using a non-Markovian expert below. 

\begin{figure}[h!]
    \centering
    \includegraphics[width=1\linewidth]{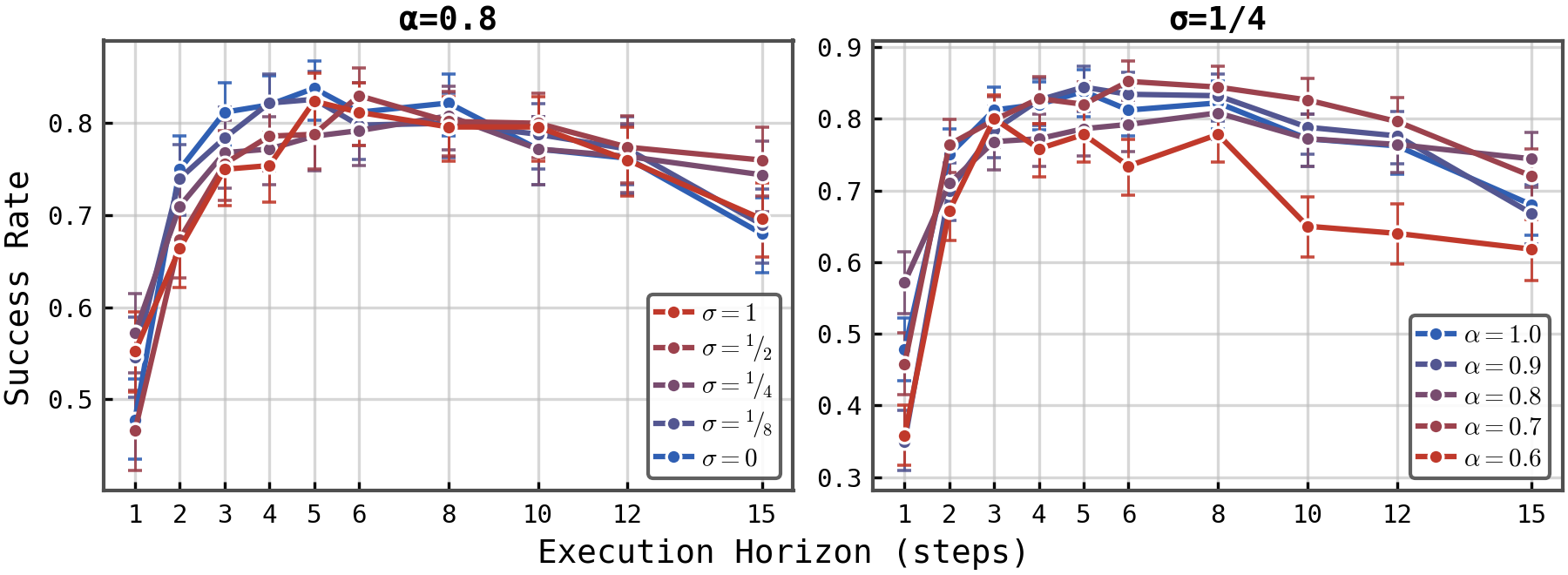}
    \caption{\textbf{Success-horizon curves of policies trained on scripted non-Markovian expert data under exploratory noise injection.} \textbf{Left}: ablation over the action-noise scale $\sigma_u$, with fraction of clean trajectories fixed at $\alpha=0.8$. \textbf{Right}: ablation over clean-trajectory fraction $\alpha$, with $\sigma_u=\frac{1}{4}$. We use $T_o=1$ in this experiment for reasons described below.}
    \label{fig:noise_injection_ablation}
\end{figure}

We sweep $\sigma_u \in \{\frac{1}{16}, \frac{1}{8}, \frac{1}{4}, \frac{1}{2}, 1\}$ and $\alpha \in \{0.6, 0.7, 0.8, 0.9\}$ and show results in \cref{fig:noise_injection_ablation}.

Surprisingly, moderate $\alpha$ and $\sigma_u$ values improve success rates at both very low $T_{\text{exec}}=1$ and at very high $T_{\text{exec}}=8,10,12,15$. Further, poor selection of $\alpha$ and $\sigma_u$ degrades performance: small $\alpha$ leaves too few clean trajectories, while insufficient or excessive action noise $\sigma_u$ provides less useful corrective supervision. The experiment doesn't show any clear shift of the success-horizon curve preferring lower execution horizons, but we also discuss caveats with this result in the coming paragraphs.

First, because we use a non-Markovian expert in this experiment, we must use a single-frame context $T_o=1$ and remove proprioceptive velocity from the observation space. Velocity and multi-frame observations typically provide strong signals of the expert's intended direction of motion. Under noise injection, however, these signals also encode the applied perturbations, which can obscure the expert's latent intent and conflate reduced observability with compounding-error mitigation. This caveat does not bias the result, but explains why we use $T_o=1$ in this experiment.

Second, even with a single-frame context, noise injection may increase multimodality. Our non-Markovian scripted expert uses an FSM with latched state transitions: if a perturbation moves the system back across a transition boundary immediately after the FSM advances, the FSM remains in its new state. Similar physical states may therefore receive actions associated with different FSM states, introducing additional target ambiguity. Because the applied perturbations are small relative to the expert's commanded actions (i.e. 0.005 m standard deviation noise at the highest $\sigma_u=1$ versus 0.025 m delta actions), we expect this effect to be limited, but it remains relevant when interpreting \cref{fig:noise_injection_ablation}.

Third, the state-action-sequence mismatch described in \cref{fn:noise-injection-chunking-caveat} may disproportionately affect policies with long execution horizons. Although all policies predict multi-step action chunks, long-execution-horizon policies rely more heavily on later actions in each chunk, where this mismatch is greatest. This artifact therefore biases the comparison toward preferring shorter execution horizons.

These caveats introduce opposing biases: additional multimodality may favor long execution horizons, whereas state-action-sequence mismatch may favor short execution horizons. We therefore do not treat this experiment as a clean test of whether exploratory noise injection reduces the need for long open-loop execution. Nevertheless, it provides a direct comparison with a complementary compounding-error mitigation technique proposed by \citet{zhang2025actionchunkingexploratorydata}. Within this setting, exploratory noise injection improves absolute performance for some hyperparameter choices, but provides no clear evidence of systematically reshaping the success-horizon curve toward shorter execution horizons.

\section{Additional Details on Long-Context Training}

\subsection{Double Encoder Method Details}
\label{sec:appendix_long_context_training_method}

We use a U-Net architecture with a double encoder and cross-attention conditioning. The U-Net and cross-attention conditioning architecture implementations are taken from \cite{agarwal2026trainingevaluatingdiffusionpolicies}. 

The double encoder is original work. It instantiates two \emph{architecturally identical but independently parameterized} observation encoders: a long-range encoder $E_{\text{L}}$ and a short-range encoder $E_{\text{S}}$. The long-range encoder is
applied to all $T_o$ frames, producing long-range tokens $\mathbf{z}^{\text{L}}_{t} = E_{\text{L}}(\mathbf{o}_t)$ for $t = 0, \dots, T_o-1$. The short-range encoder is applied only to the most recent $T_s$
frames ($T_s \le T_o$), producing short-range tokens $\mathbf{z}^{\text{S}}_{t} = E_{\text{S}}(\mathbf{o}_t)$ for $t = T_o - T_s, \dots, T_o-1$. The two streams are concatenated into a single conditioning sequence of $T_o + T_s$ tokens that the denoising network conditions on via cross-attention. Consequently, the most recent $T_s$ frames are represented \emph{twice} --- once through the shared long-horizon pathway and once through the short-horizon pathway. This allows the denoising network to attend to a high-capacity, specialized representation of the immediate past while still retaining the full temporal context. Each conditioning token receives two additive embeddings before entering the cross-attention blocks: (i) a \emph{temporal} embedding, a fixed sinusoidal positional encoding indexed by the token's original timestep index; and (ii) a learned \emph{range} embedding that tags each token as \textsc{long} or \textsc{short}, letting the network distinguish the two pathways (a reserved \textsc{null} entry is used for the prepended diffusion-timestep token). All tokens are linearly projected to a shared width before being attended to at every residual block of the 1D U-Net denoiser via multi-head cross-attention. To prevent the policy from over-relying on the short-horizon pathway, we apply \emph{short-range dropout} during training: for each sample in a minibatch, with probability $p_{\text{short\_range\_dropout}}$ all $T_s$ short-range tokens of that sample are replaced by a single learned null token $\mathbf{z}_{\varnothing} \in
\mathbb{R}^{d}$. This dropout is disabled at inference time. In our experiments we use $T_s = 2$ and $p_{\text{short\_range\_dropout}} = 0.3$.

Note that we use the \textit{past-token-prediction} auxiliary training objective from \cite{chi2024diffusionpolicyvisuomotorpolicy,torne2025learninglongcontextdiffusionpolicies}, where the policy's predicted action sequence includes actions corresponding to every observation in the observation sequence. This means $T_o - 1$ predicted actions are in the past. In this paper, we treat the prediction horizon $T_p$ as referring to only the predicted actions in the future and fix this at 15 timesteps, but the actual sequence the policy learns to predict has length $T_p + T_o - 1$.

\subsection{Double Encoder Method Analysis}
\label{sec:appendix_long_context_training_analysis}

\begin{figure}[h!]
    \centering
    \includegraphics[width=0.925\linewidth]{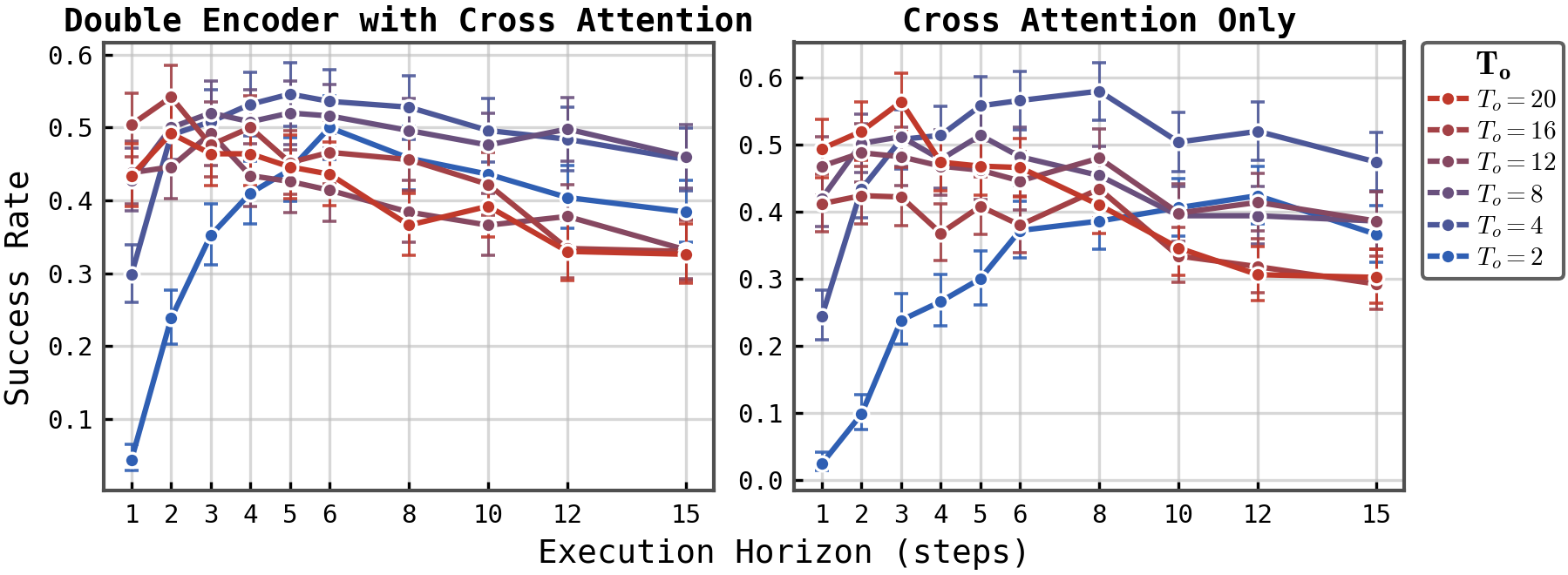}
    \caption{\textbf{\texttt{FurnitureSimOneLeg} success-horizon curves for $T_o \in \{2, 4, 8, 12, 16, 20\}$ for policies trained with double encoder cross attention U-Net and cross attention U-Net (no double encoder) architectures on 200 human teleop demonstrations.} Both plots exhibit similar trends (shorter $T_{\text{exec}}$ perform better as $T_o$ increases) and comparable absolute performance, but the double encoder architecture exhibits much less variation in performance across execution horizons, producing smoother and more interpretable curves.}
    \label{fig:double_encoder_vs_no_double_encoder}
\end{figure}

In this section, we provide a brief comparison between long-context policies (following the cross-attention U-Net architecture from \cite{agarwal2026trainingevaluatingdiffusionpolicies}) trained with and without the double encoder on the \texttt{FurnitureSimOneLeg} task. See \cref{fig:double_encoder_vs_no_double_encoder} for the results.

While both yield similar absolute performance and exhibit the same trend from \cref{sec:Long Context Provides an Alternative to Action Chunking}, where longer context lengths improve relative performance at low execution horizons (further supporting that our results are architecture-agnostic), the policy with the double encoder appears to have much less performance variation between execution horizons. Although the mechanism underlying this effect is unclear, we hypothesize that the double encoder may be providing redundant representations of short-range observations, reducing brittleness through either pathway. For this reason, we use the double encoder in most of our long context experiments.

\subsection{Context Length Ablations for \texttt{PushT-M}}
\label{sec:pusht-m-additional-discussion}

\begin{figure}[h!]
    \centering
    \includegraphics[width=0.475\linewidth]{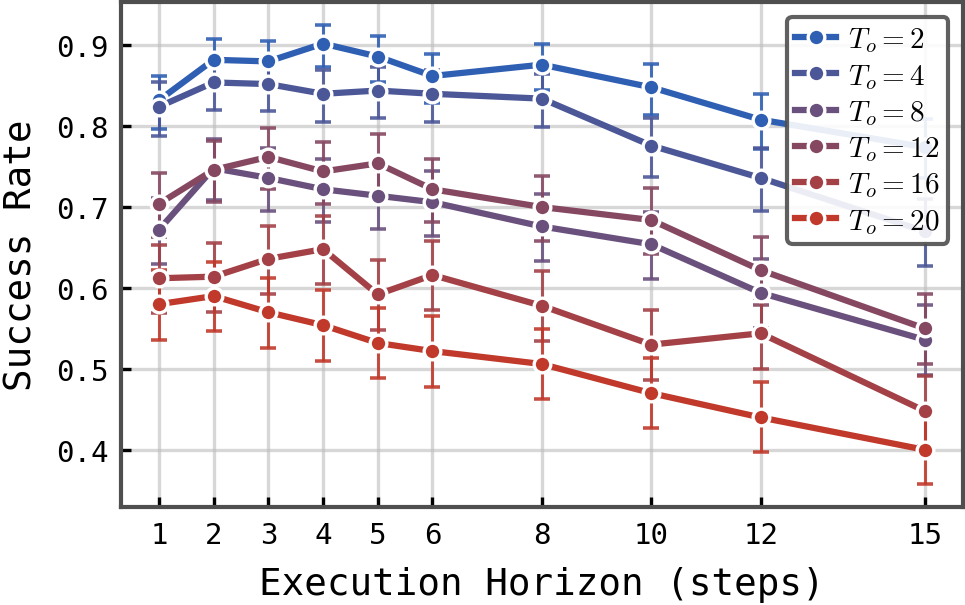}
    \caption{\textbf{Success-horizon curves from long-context policies trained in \texttt{PushT-M}.}}
    \label{fig:context_length_ablation_pusht_m_appendix}
\end{figure}

While our main context-length ablations in
\cref{sec:Long Context Provides an Alternative to Action Chunking} use \texttt{PushT-D}, for completeness, we also train and evaluate long-context policies in \texttt{PushT-M}. Results are shown in
\cref{fig:context_length_ablation_pusht_m_appendix}.

As in \texttt{PushT-D} (\cref{fig:context_length_ablation_3_tasks}), increasing context length shifts performance toward shorter execution horizons. However, unlike in our other tasks, absolute performance in \texttt{PushT-M} degrades substantially as context length increases. We hypothesize that this degradation reflects causal confusion exacerbated by visually sparse rendering in the ManiSkill simulator used in \texttt{PushT-M} --- the lighting is bright and there are few visual features, causing unrelated states to appear similar in the learned representation (we observe that in some rollouts, the policy appears to mistake the T’s upper corners for its base). This additional state aliasing may exacerbate causal confusion when conditioning on longer histories. In all other tasks, we find that multiple long-context training methods readily work out of the box. Because long-context training is not the focus of this work and this degradation may be specific to \texttt{PushT-M}'s simulation rendering, we leave it as future work to investigate further.

A separate observation is that the $T_o=2$ curve in \cref{fig:context_length_ablation_pusht_m_appendix} can be compared directly with the human-expert \texttt{PushT} curve in \cref{fig:human_vs_markovian_3_tasks}. Both experiments use the same environment, demonstrations, and context length; only the architecture differs: the results here use the double encoder architecture, versus a FiLM-conditioned U-Net in \cref{fig:human_vs_markovian_3_tasks}. Both produce the same trend: an inverted U-shaped success-horizon curve with $T_{\text{exec}}^*$ of 4 or 5. However, we notice an interesting phenomenon: the double encoder policy achieves substantially higher performance at $T_{\text{exec}}=1$, suggesting architecture may also influence the success-horizon curve. 

To explain this, we speculate that the double encoder’s redundant representations and short-range dropout improve robustness to policy-induced states, thereby reducing compounding errors at low $T_{\text{exec}}$. Nevertheless, the improved architecture only changes the magnitude of the low-horizon performance deficit, not the qualitative preference: short-context policies imitating non-Markovian experts still prefer a non-trivial $T_{\text{exec}}$ of 4 -- 5. Meanwhile, increasing context length to improve observability of non-Markovian behaviors still produces a much more substantial shift of $T_{\text{exec}}^*$ to 1 -- 2. Thus, while architectural changes that reduce compounding errors may improve low-horizon performance, their influence on the success-horizon curve appears much weaker than changes in expert non-Markovianity.

\section{Additional Discussion: Multimodality and Smooth Execution}
\label{appx:multimodality_and_smooth_execution}

There is a subtle interplay between multimodality in expert demonstrations and smooth execution. Let us assume that every expert demonstration trajectory is smooth,\footnote{Here, ``smoothness" refers qualitatively to discrete-time smoothness of the action sequence, i.e., actions and their higher-order discrete derivatives vary gradually over time.} but over the batch of demonstrations there are cases where the same history can lead to qualitatively different actions (multimodal), e.g. as might have been the case for Buridan's Ass. Since the learned policy mimics the \emph{distribution} of actions, the concept of smoothness coming from each individual trajectory might be lost, and one might worry again about the learned policy cycling between two different ``modes" in the demonstration data.

Let us assume for the moment that the learner learns perfectly (reconstructing the true distribution optimally given the available context). If the history/context of the learner is the same or greater than the expert, then all sampled trajectories must also be smooth (non-smooth trajectories would have zero probability mass). The more interesting case is where the learner has a more limited context than the expert. Smoothness is only a local property of the trajectories and does not require arbitrarily long histories. Technically, we are working with sampled-data systems here, so we can only talk about ``discrete smoothness" (e.g., high-order differences quickly decay); if the history contains $k$ recent samples at the execution sampling rate, then we expect that the outputs will be discretely smooth to order $k-1$.

Thus, under perfect learning, we argue that expert multimodality should not by itself cause arbitrarily unsmooth execution. Learning errors (including due to finite demonstration data) can complicate this, but we leave that analysis to future work.

\section{Additional Discussion: Benefits of Reactivity in Our Experiments}
\label{sec:appendix_reactivity}

In this section, we provide more concrete examples of the benefits of reactivity in our experiments.

We primarily attribute its benefits to challenging contact dynamics (particularly in \texttt{FurnitureSimOneLeg}, during the screwing phase, and in \texttt{Push-T}) and high precision requirements (particularly in \texttt{FurnitureSimOneLeg} and \texttt{GearInsertion}). In \texttt{FurnitureSimOneLeg}, at high $T_{\text{exec}}$, we find a vast majority of failures occur during insertion (by missing the socket) and screwing (due to the table top sliding away); these failures become very infrequent at low $T_{\text{exec}}$ where the policy can re-aim and adjust when the table begins to slide. Most high-$T_{\text{exec}}$ failures in \texttt{GearInsertion} occur due to missed insertion; in \texttt{Push-T} --- due to unexpected T motion that require additional pushing maneuvers (leading to timeout); in \texttt{Kitchen} --- due to lack of precision when grasping objects.

These failure modes illustrate that even nominally static manipulation tasks suffer inevitable execution errors, from imperfect policy learning, control, interaction dynamics, or otherwise. Reactivity allows policies to quickly correct such errors, providing beneficial robustness even in otherwise static settings.

\section{Additional Discussion: The Importance of Checkpoint Selection for Long-Context, Reactive Policies}
\label{sec:appendix_checkpoint_selection}

Across almost all experiments, we noticed two consistent trends. While conventional Diffusion Policy wisdom promotes over-training past the epoch of minimum validation loss \cite{he2025demystifyingdiffusionpoliciesaction} to achieve strong rollout performance, we find:

\begin{enumerate}
    \item \textbf{Lower execution horizons prefer earlier checkpoints, i.e. less over-training.}
    \item \textbf{Longer context lengths also prefer earlier checkpoints.}
\end{enumerate}

Claim 1 holds fairly consistently across all experiments. We find claim 2 less consistent --- we observe it strongly across the \texttt{FurnitureSimOneLeg}, \texttt{SinglePillDispense}, and \texttt{SlipIntoBaggie} experiments; in \texttt{SinglePillDispense}, naively evaluating the latest checkpoint even yields up to 40\% lower success rates compared to an earlier checkpoint. But we observe no pattern at all in \texttt{Push-T}, \texttt{GearInsertion}, and \texttt{Kitchen}. We also acknowledge that these comparisons conflate other variables (i.e. network architecture), but nonetheless motivate evaluating checkpoints throughout training --- a precaution that may be crucial in some settings but less so in others.

\begin{figure}[h!]
    \centering
    \includegraphics[width=1\linewidth]{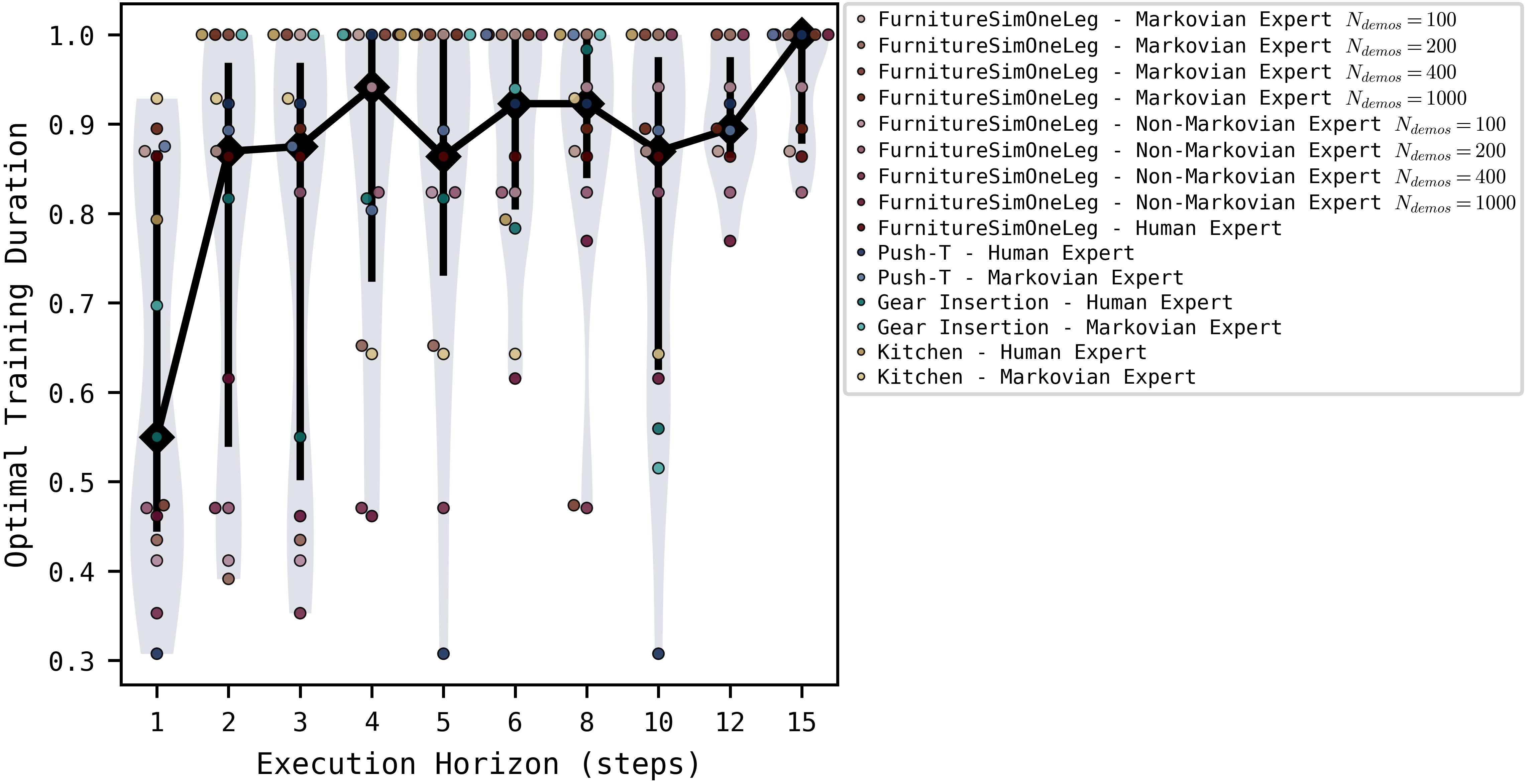}
    \caption{\textbf{Distribution of the optimal training duration (calculated as the fraction of the highest-performing checkpoint's epoch divided by the latest evaluated checkpoint's epoch) as a function of execution horizon}, pooled across all short-context ($T_o=2$) scripted- and human-expert experiments (individual points represent individual experiments; black diamonds/bars represent median and interquartile range). We observe a consistent trend across almost all experiments that lower execution horizons prefer earlier checkpoints.}
    \label{fig:checkpoint_selection}
\end{figure}

\begin{figure}[h!]
    \centering
    \includegraphics[width=1\linewidth]{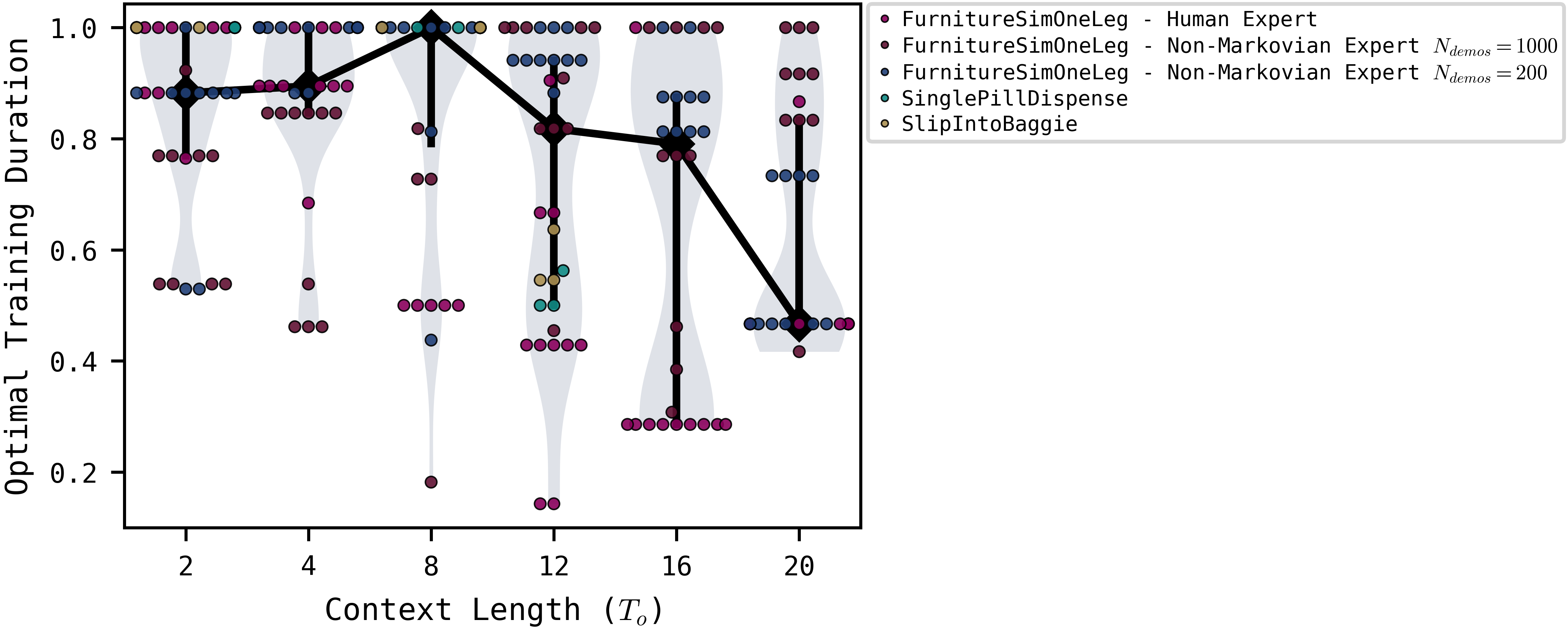}
    \caption{\textbf{Distribution of the optimal training duration as a function of context length} for all \texttt{FurnitureSimOneLeg}, \texttt{SinglePillDispense}, and \texttt{SlipIntoBaggie} context length experiments (including all execution horizons). We observe policies with longer context lengths often perform better at earlier training checkpoints.}
    \label{fig:checkpoint_selection_context_length}
\end{figure}

We plot the optimal training duration against execution horizon in \cref{fig:checkpoint_selection}, and against context length in \cref{fig:checkpoint_selection_context_length}.

One possible explanation for this phenomenon is that early-stopping training reduces overfitting and improves generalization to unseen states \cite{raskutti2013earlystoppingnonparametricregression}, which is especially important when using long context lengths or short execution horizons --- long context lengths expand the policy’s input space, increasing risk of encountering out-of-distribution inputs during rollout (see \cref{sec:Long Context Provides an Alternative to Action Chunking}), while short execution horizons allow errors on policy-induced out-of-distribution states to compound more readily (see \cref{sec:Investigating the Role of Compounding Errors}). \textbf{The implication is that, to effectively train policies with either long context lengths and/or short execution horizons, adopting careful checkpoint selection processes is critical.}

This implication may become even more important at large data scales. Results from Generalist's GEN-0 and Sunday's ACT-2 suggest that, with enough data, validation loss becomes a strong predictor of closed-loop performance \cite{generalist2025gen0,sunday2026act2preview}. In this regime, policies no longer face the same tradeoff between generalization and rollout performance: they can generalize well \textit{and} perform strongly without over-training. We therefore hypothesize that long-context, reactive policies will become even more favorable at scale. Nevertheless, we note that this analysis is preliminary, and much remains to be understood about the relationship between checkpoint selection, validation loss, and rollout performance in behavior cloning.

In relation to \cref{sec:Investigating the Role of Compounding Errors}, this result further contributes that compounding errors especially impact lower execution horizons; but proper checkpoint selection is a simple and standard way to limit this effect.

\end{document}